\documentclass[sigconf]{acmart}
\AtBeginDocument{%
  }

\setcopyright{cc}
\setcctype{by}
\copyrightyear{2026}
\acmYear{2026}
\acmBooktitle{Proceedings of the ACM Web Conference 2026 (WWW '26), April 13--17, 2026, Dubai, United Arab Emirates}
\acmPrice{}

\acmDOI{10.1145/3774904.3792618}
\acmISBN{979-8-4007-2307-0/2026/04}
\acmConference[WWW '26]{Proceedings of the ACM Web Conference 2026}{April 13--17, 2026}{Dubai, United Arab Emirates}

\usepackage{hyperref} 
\usepackage{multirow}
\usepackage{newfloat}
\usepackage{amsmath}

\usepackage{caption}

\usepackage{xcolor}
\usepackage{makecell}
\usepackage{algorithm}
\usepackage{algorithmic}

\begin{document}

\title[Time-TK: A Multi-Offset Temporal Interaction Framework for Time Series Forecasting]{Time-TK: A Multi-Offset Temporal Interaction Framework Combining Transformer and Kolmogorov-Arnold Networks for Time Series Forecasting}

\author{Fan Zhang}
\orcid{0000-0002-0343-3499}
\affiliation{%
	\institution{School of Computer Science and Technology, Shandong Technology and Business University}
	\city{Yantai}
	\state{ShanDong}
	\country{China}}
\email{zhangfan@sdtbu.edu.cn}

\author{Shiming Fan}
\orcid{0009-0000-4858-8315}
\affiliation{%
  \institution{School of Computer Science and Technology, Shandong Technology and Business University}
  \city{Yantai}
  \state{ShanDong}
  \country{China}}
\email{2024410061@sdtbu.edu.cn}

\author{Hua Wang}
\authornote{Corresponding author.}
\orcid{0000-0002-8844-9667}
\affiliation{%
	\institution{School of Computer and Artificial Intelligence, Ludong University}
	\city{Yantai}
	\state{ShanDong}
	\country{China}}
\email{hwa229@163.com}

%
%
%
%


\begin{abstract}
Time series forecasting is crucial for the World Wide Web and represents a core technical challenge in ensuring the stable and efficient operation of modern web services, such as intelligent transportation and website throughput. However, we have found that existing methods typically employ a strategy of embedding each time step as an independent token. This paradigm introduces a fundamental information bottleneck when processing long sequences, the root cause of which is that independent token embedding destroys a crucial structure within the sequence—what we term as multi-offset temporal correlation. This refers to the fine-grained dependencies embedded within the sequence that span across different time steps, which is especially prevalent in regular Web data. To fundamentally address this issue, we propose a new perspective on time series embedding. We provide an upper bound on the approximate reconstruction performance of token embedding, which guides our design of a concise yet effective Multi-Offset Time Embedding (MOTE) method to mitigate the performance degradation caused by standard token embedding. Furthermore, our MOTE can be integrated into various existing models and serve as a universal building block. Based on this paradigm, we further design a novel forecasting architecture named Time-TK. This architecture first utilizes a Multi-Offset Interactive KAN (MI-KAN) to learn and represent specific temporal patterns among multiple offset sub-sequences. Subsequently, it employs an efficient Multi-Offset Temporal Interaction mechanism (MOTI) to effectively capture the complex dependencies between these sub-sequences, achieving global information integration. Extensive experiments on 14 real-world benchmark datasets, covering domains such as traffic flow and BTC/USDT throughput, demonstrate that Time-TK significantly outperforms all baseline models, achieving state-of-the-art forecasting accuracy.
\end{abstract}

\ccsdesc[500]{Applied computing~Forecasting}
\ccsdesc[300]{Computing methodologies~Information extraction}
\ccsdesc[300]{Information systems~Traffic analysis}
%

\keywords{Kolmogorov-Arnold Networks, Web time series data, Multi-Offset embedding mechanism}


\maketitle

\section{Introduction}

\begin{figure}[t]
	\centering
	\includegraphics[width=0.48\textwidth]{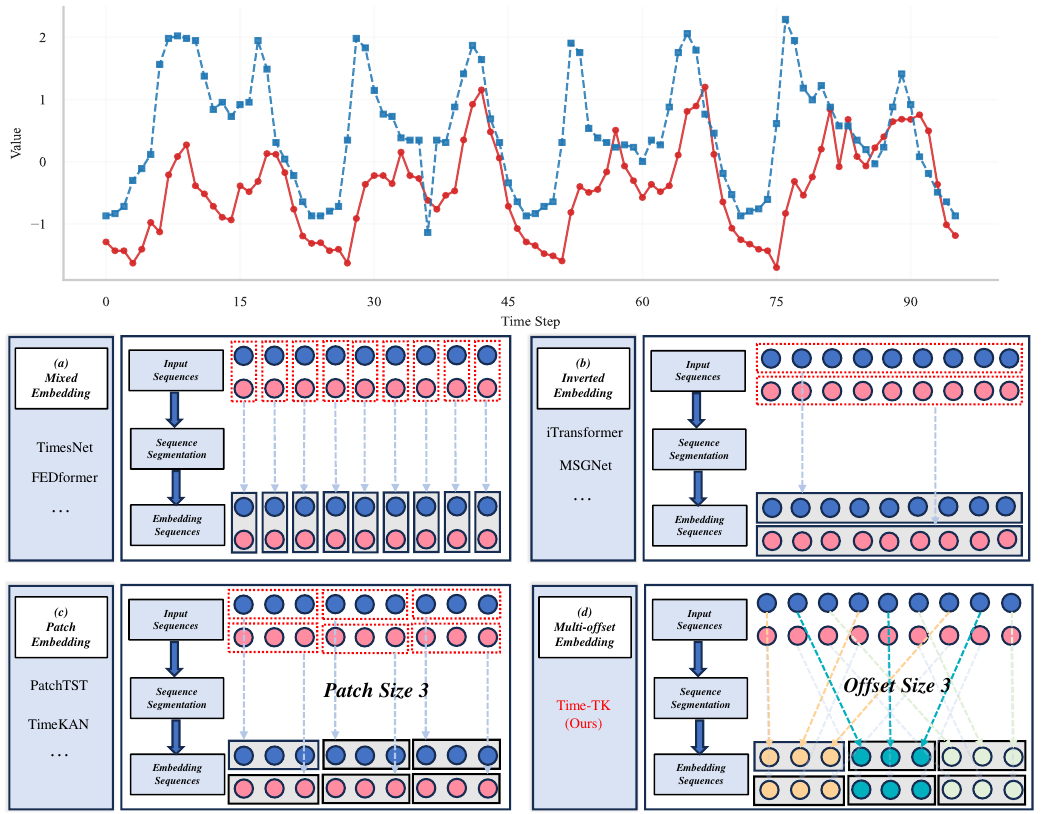} 
	\caption{Illustration of four time series embedding strategies. (a) Mixed embedding of variables at the same time step. (b) Inverted embedding along the time axis. (c) Patch embedding based on temporal segmentation. (d) Multi-Offset embedding mechanism used in the proposed Time-TK.}
	\label{fig1}
\end{figure}

%

In the vast and dynamic ecosystem of the World Wide Web, Long-Term Time Series Forecasting (LTSF) has emerged as a crucial research frontier \citep{cai2025learning,huang2025exploiting,1,ke2025early,liu2026healthwealth,shen2025aienhanced}. Web platforms themselves are generators of massive time-series data \citep{teng2025timechain,xie2025multivariate}, ranging from website traffic and user engagement metrics to the continuous data streams produced by Internet of Things (IoT) devices interacting via APIs. Unlike information-dense data such as images or text, the core value of time-series data lies in its temporal dynamics rather than in isolated, individual time points \citep{fan2025cawformer,qiu2024duet}. This information sparsity at a single time step makes it extremely challenging to extract meaningful patterns from the data \citep{alkhanbouli2025role}. Consequently, the research focus has definitively shifted towards modeling complex temporal dynamics. This shift not only profoundly aligns with the intrinsic characteristics of web-generated data but also serves as the foundation for uncovering key underlying structures, such as the periodicity of daily user activity and long-term platform growth (trends), thereby enabling proactive resource planning and intelligent web services.

Recently, Transformer-based models have received increasing attention in the context of long-term time series forecasting. However, some work has shown that Transformer exhibits suboptimal performance in long-term multivariate time series forecasting tasks\citep{10}. This is mainly attributed to the fact that most existing LTSF methods focus on reducing computational costs in univariate settings and lack targeted modeling approaches that address the unique characteristics of long-term multivariate time series. Time series data typically exhibit strong periodic characteristics along with irregular fluctuations, as illustrated in Figure \ref{fig1}. Therefore, the token embedding approach based on a single time step \citep{8,24,25} is difficult to capture these key features effectively. Some studies have attempted to perform token embedding in the time dimension \citep{13,30} to enhance the model's understanding of the periodic structure. However, these methods typically rely on holistic sequence embeddings, which may overlook fine-grained temporal dynamics essential for accurate forecasting \citep{18}. In addition, LTSF is more prone to overfitting than short-term prediction tasks. During training, the model may overfit to noisy patterns in the data and fail to capture the underlying trends and true temporal dependencies.


To address these issues, we propose Time-TK, a multi-offset temporal interaction framework that integrates Transformer and KAN. It focuses on capturing deep temporal dependencies from historical time series to enhance the model’s long-term forecasting capability. Given that long-term time series forecasting relies heavily on modeling extensive historical information, we design our approach around the inherent temporal structure of the data. Specifically, we introduce a multi-offset temporal token embedding mechanism, as shown in Figure \ref{fig1}, which divides the original time series into multiple sub-sequences with different spans at fixed offsets along the temporal dimension and performs independent embedding operations on each sub-sequence. The Multi-Offset Interactive KAN (MI-KAN) module leverages the flexibility of KAN \citep{liu2024kan,bresson2024kagnns} in kernel function modeling to deeply model the temporal structure within each offset sub-sequence and capture its unique dynamic patterns. Based on these offset embedding tokens, the multi-offset temporal interaction module captures cross-step dependencies between time steps and compensates for long-term interleaved dynamics that are often overlooked by traditional continuous embedding methods. To achieve a more comprehensive understanding of the time series, we further design a global interaction mechanism that jointly encodes the original sequence with the offset sub-sequences. This helps recover missing information in cross-offset segments and integrates it into a unified global representation, enhancing the model’s ability to capture long-term global structure. As shown in Figure \ref{fig:2}, Time-TK achieves state-of-the-art performance on several long-term time series forecasting tasks. It also adopts a lightweight architecture that outperforms more complex TSF models while using fewer computational resources.

Our main contributions are as follows:


\begin{figure}[t]    
	\centering
	\begin{minipage}[b]{0.48\textwidth} 
		\includegraphics[width=\textwidth]{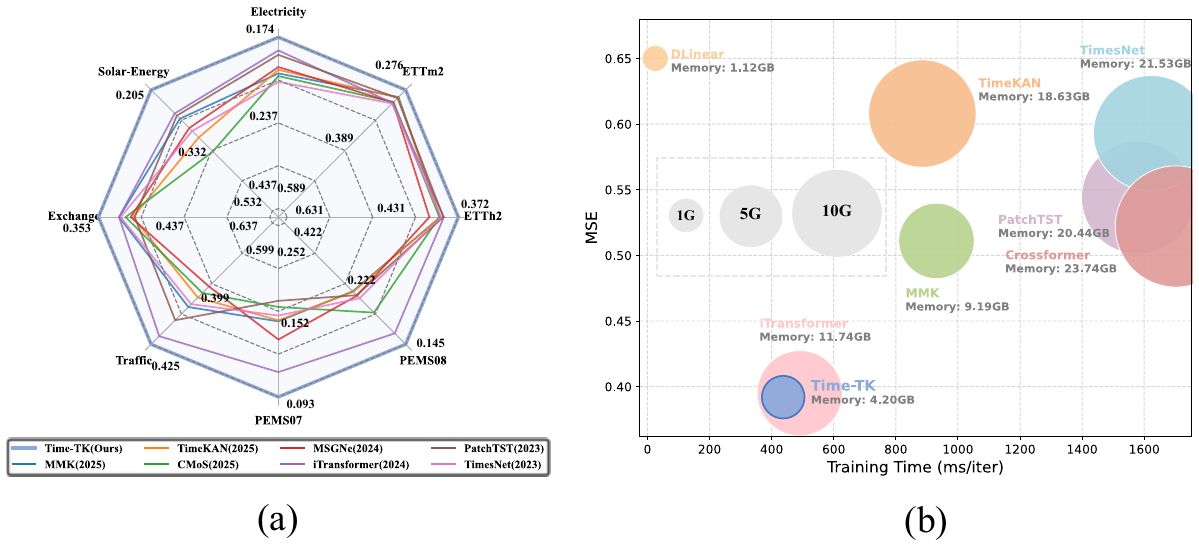}
		\centering
	\end{minipage}
	\caption{(a) Average performance across all prediction windows, showing improvements over the baseline on various datasets. (b) Comparison of memory usage (GB), training time (ms/iter), and MSE on the Traffic dataset. The prediction length was set to 96.}
	\label{fig:2}
\end{figure}

\begin{itemize}
	\item We find that existing embedding methods cannot effectively capture the dependencies between different time steps. To address this problem, this paper proposes a multi-offset temporal token embedding method, which is one of the few ways to explore directly from the original sequence.
	\item Time-TK is a lightweight and efficient model that incorporates the MI-KAN module. Leveraging the flexibility of KAN, it effectively models multi-offset sub-sequences. Moreover, Time-TK is among the few time series forecasting models that successfully integrate Transformer and KAN.
	\item  We conduct extensive experiments on 14 real-world datasets,  and the results demonstrate that Time-TK consistently achie-ves state-of-the-art performance, validating its effectiveness for long-term time series forecasting.
\end{itemize}
\begin{figure*}[ht]    
	\centering
	\begin{minipage}[b]{0.94\textwidth} 
		\includegraphics[width=\textwidth]{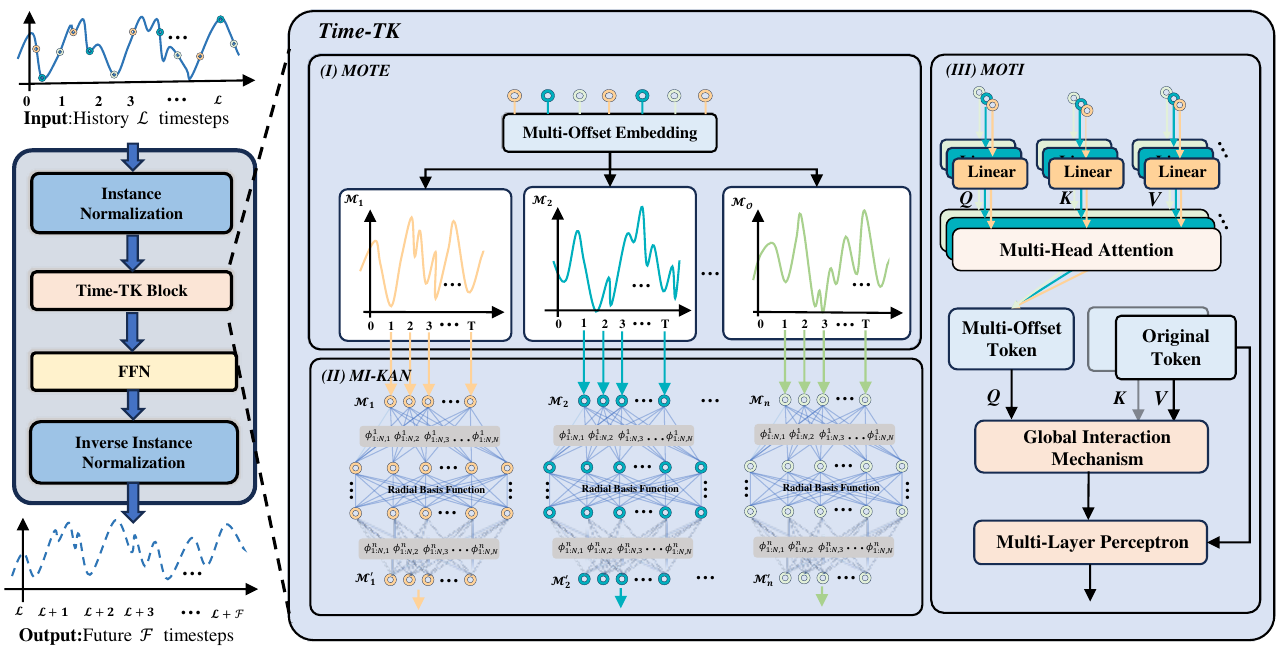}
		\centering
	\end{minipage}
	\caption{Overall architecture of Time-TK. MOTE performs Multi-Offset token embedding on the sequence, followed by MI-KAN learning representation of the subsequences, and finally interactive prediction through MOTI.}
	\label{fig:model}
\end{figure*}

\section{Related Works}
With the breakthroughs of deep learning \citep{SEExiao,lin2025cec,chen2024confusion,xiao2025curiosity,li2025multi,li2023ultrare,xiao2025points,yao2024swift} in natural language processing \citep{11270220,chen2024post} and computer vision \citep{zhang2024cf,zhang2023multi}, its application in time series forecasting has also grown rapidly. Traditional methods such as ARIMA \citep{ariyo2014stock} are constrained by linear assumptions, making them inadequate for capturing nonlinear dynamics in temporal data. In contrast, deep learning models such as RNNs \citep{hochreiter1997long,ke2025stable}, LSTMs \citep{gers2000learning}, and Transformers \citep{vaswani2017attention} have significantly improved forecasting accuracy by learning time dependencies. Embedding strategies play a crucial role in time series modeling, as they transform low-dimensional raw inputs into high-dimensional representations, helping models to capture underlying temporal structures and semantic patterns. In this section, we summarize the mainstream embedding approaches for time series. As shown in Figure 1, these methods can generally be divided into three categories:The first category \citep{8,28,9,25} employs channel-mixing mechanisms, in which each timestep is represented by the integration of cross-channel latent features. However, MLP-based models \citep{10} have raised the question: “Are Transformers effective for time series forecasting?” With their outstanding performance and efficiency, they pose a significant challenge to the effectiveness of such Transformer-based methods. The second category \citep{16} adopts patch-based embeddings by segmenting the sequence into local windows to preserve segment-level semantics, thereby capturing broader temporal patterns that are often missed by pointwise models. The third category \citep{15} introduces an inverted embedding mechanism, where complete sub-sequences along the temporal axis are embedded into single tokens, allowing each token to aggregate global sequence representations. This design aligns well with attention-based architectures and has received considerable attention.

Unlike the aforementioned strategies, we explore a novel embedding mechanism aimed at enhancing the model's ability to learn specific temporal patterns. Our method demonstrates consistently effective performance across a variety of experimental settings, validating its applicability to time series forecasting tasks.
\section{Methodology}

\subsection{Overview of Time-TK}
Given a historical time series \(\mathcal{X} = [{x_1},... ,{x_\mathcal{L}}] \in {{\mathbb R}^{N \times \mathcal{L}}}\), the objective of time series forecasting is to predict future values \({\mathcal{\hat{Y}}_t} = [{x_{\mathcal{L}+1}},... ,{x_{\mathcal{L} + \mathcal{F}}}] \in {{\mathbb R}^{N \times \mathcal{F}}}\), where \(N\) is the number of variables, \(\mathcal{L}\) is the length of the input sequence, and \(\mathcal{F}\) is the forecast horizon. As shown in Figure \ref{fig:model}, our proposed Time-TK architecture consists of multiple stages. First, Multi-Offset Token Embedding divides the original sequence into multiple sub-sequences with different time spans. MI-KAN (Multi-Offset Interactive KAN) learns and represents specific temporal patterns between offset sub-sequences. The Multi-Offset Temporal Interactive module captures long distance dependencies across time steps based on the representation of these offset sub-sequences. At a higher level, the global interaction mechanism further fuses the contextual information of the original sequence with that of all offset sub-sequences to capture the missing information across offset segments and unify it into the global representation. The final prediction is obtained by mapping the learned representation through a linear projection layer. The core modules of Time-TK are introduced in detail below, while the complete algorithmic workflow is provided in Appendix \ref{apC}. The code is available from the repository\footnote{\url{https://github.com/fsmss/Time-TK}}.

\subsection{Multi-Offset Token Embedding}
Three main forms of existing embedding methods exist: (i) uses embedding based on a single time step also called channel mixing (CM), (ii) takes the entire time dimension as the embedding input. (iii) segments the time dimension for embedding. However, these approaches often struggle to adequately capture dependencies across different time scales, especially in periodic and non-stationary time series, leading to limited interaction capabilities. To address this limitation, we propose a Multi-Offset Token Embedding strategy. Specifically, given a predefined offset size \(\mathcal{O}\), we divide the historical sequence into multiple sub-sequences \(\{ {\mathcal{M}_1},...,{M_{\mathcal{O}}}\} \). As shown in Figure~\ref{fig1}, unlike traditional approaches, we use multiple sub-sequences with different temporal offsets as token inputs to capture information across varying time scales. This design enables the model to capture time-dependent features at different granularities in long sequences—for example, some sub-sequences are more effective at modeling short-term fluctuations, while others are better suited for long-term trends. The introduction of Multi-Offset Token Embedding significantly enhances the model's adaptability to complex temporal patterns, while effectively mitigating overfitting caused by noise in the training data, thereby improving overall generalization.

\subsection{Multi-Offset Interactive KAN}
After the Multi-Offset Token Embedding process, we obtain multiple offset sub-sequences \(\{ {\mathcal{M}_1},...,{M_{\mathcal{O}}}\} \). To further capture the temporal dependencies both within and across these sub-sequences, we design the Multi-Offset Interactive KAN (MI-KAN) module, which aims to learn dedicated representations for each offset sub-sequence and model their mutual relationships. Compared with traditional MLPs, KAN (Kolmogorov-Arnold Network)\citep{liu2024kan} focuses more on approximating complex, high-dimensional mapping relationships through a set of combinable simple functions. Specifically, KAN enhances the network's ability to model nonlinear patterns by replacing traditional linear connections between neurons with learnable univariate functions. The mapping between neurons in adjacent layers can be formulated as:
\begin{equation}\label{eq:1}
	\begin{array}{l}
		\mathcal{Z}_j^{(l + 1)} = \sum\limits_i {{\phi _{ij}}(\mathcal{Z}_i^{(l)})} 
	\end{array}
\end{equation}
Where \({\mathcal{Z}_i^{(l)}}\) represents the \(i\)th neuron in the \(l\)th layer, \({\mathcal{Z}_j^{(l+1)}}\) represents the \(j\)th neuron in the \((l+1)\)th layer, and \({{\phi _{ij}}}\) is the univariate mapping function from the \(i\)th to the \(l\)th neuron. Early KAN implementations usually used spline functions as basic building blocks, but such methods often require complex rescaling and have poor stability when dealing with variables crossing the boundaries of the domain. To address these limitations, we adopt the more efficient and stable FastKANLayer \citep{li2024kolmogorov}, which constructs univariate mappings using combinations of radially symmetric functions and offers greater flexibility and generalization capability. In our implementation, we employ Gaussian radial basis functions (RBFs) to model the nonlinear relationships in the input. The RBF is defined as follows:
\begin{equation}\label{eq:2}
	\begin{array}{l}
		\phi (r) = \exp ( - \frac{{{r^2}}}{{2{h^2}}})
	\end{array}
\end{equation}
Where \(r\) represents the distance between the input and the center, and \(h\) controls the smoothness of the function. The output of the RBF network is a linear combination of this radial basis function, weighted by an adjustable coefficient. The output of the entire RBF network can be expressed as:
\begin{equation}\label{eq:3}
	\begin{array}{l}
		f(x) = \sum\limits_{i = 1}^N {{w_i}} \phi (\left\| {x - {x_i}} \right\|)
	\end{array}
\end{equation}
Where \(w_i\) is the learnable weight and \(x_i\) is the RBF center. It is worth noting that FastKANLayer exhibits strong expressiveness and effectively captures the complex dynamic patterns present in time series data. It generates corresponding deep representations for each input sub-sequence. This design not only simplifies the model structure but also enhances representation consistency across different sub-sequences through a unified modeling approach, thereby facilitating the interaction module in capturing correlations across time offsets. Finally, the learning process of the proposed MI-KAN module can be formulated as:
\begin{equation}\label{eq:4}
	\begin{array}{l}
		\{ {\mathcal{M}'_1},...,{M'_{\mathcal{O}}}\}  = MI-KAN({\mathcal{M}_1},...,{M_{\mathcal{O}}})
	\end{array}
\end{equation}
The offset sub-sequence representations \(\{ {\mathcal{M}'_1},...,{M'_{\mathcal{O}}} \}\in {{\mathbb R}^{\mathcal{O} \times N \times T}} \) obtained from the MI-KAN module, preserve the temporal dynamics within each individual sub-sequence.

\subsection{Multi-Offset Temporal Interaction Forecasting}	
To further capture correlations across different time steps, we introduce the Multi-Offset Temporal Interaction Mechanism. The primary objective of this mechanism is to leverage the previously proposed Multi-Offset Token Embedding to enhance the model's ability to capture implicit temporal structures across multiple related sub-sequences. Specifically, for each sub-sequences \({{\mathcal{M}'}_u}\), we apply a multi-head Self-Attention mechanism (\(MSA\)) on all its feature dimensions:
\begin{equation}\label{eq:5}
	\begin{array}{l}
		{\mathcal{A}_u} = {{\mathcal{M}'}_u} + MSA({{\mathcal{M}'}_u},{{\mathcal{M}'}_u},{{\mathcal{M}'}_u})
	\end{array}
\end{equation}
Where \({{\mathcal{M}'}_u} \in {{\mathbb R}^{N \times T}}\) represents the representation of the \(u\)-th offset sub-sequence, and \(MSA( \cdot )\) is a multi-head self-attention operation. Due to the use of piecewise offset embeddings, each sub-sequence is significantly shortened, resulting in an attention computation with approximately linear time complexity at this stage \citep{xu2024sst}. After modeling the internal structure of each sub-sequence, we further introduce a global fusion operation to integrate the interaction results of the original sequence representation \(\mathcal{X}\) with all offset sub-sequences \(\mathcal{A}\), in order to capture potential dependencies across different temporal segments. The fusion process is formally defined as:
\begin{equation}\label{eq:6}
	\begin{array}{l}
		\mathcal{H} = \mathcal{X}+MSA(\mathcal{Q}=\mathcal{A},\mathcal{K}=\mathcal{X},\mathcal{V} = \mathcal{X})
	\end{array}
\end{equation}
The sequence processed by the Multi-Offset Interaction Mechanism serves as the query, while the original sequence acts as both the key and value, enabling information fusion across different temporal offsets. To generate the final prediction result, we map the time dimension to the prediction length of the target through a linear layer. The transformation can be expressed as:
\begin{equation}\label{eq:7}
	\begin{array}{l}
		\mathcal{Y} = Linear(\mathcal{H}) \in {{\mathbb R}^{N \times \mathcal{F}}}
	\end{array}
\end{equation}
\begin{table*}[htbp]
	\centering
	\small
	\setlength{\tabcolsep}{18pt}
	\renewcommand{\arraystretch}{1.}
	\caption{Detailed description of the dataset. Dim indicates the number of variables in each dataset. Dataset Size indicates the total number of time points in (training set, validation set, test set). Prediction Length indicates the future time points that need to be predicted. Each dataset contains four prediction settings. Frequency indicates the sampling interval of the time points. Data statistics are from iTransformer \citep{15}.}
	\begin{tabular}{l|c|c|c|c|c}
		\hline
		\makecell[l]{\textbf{Dataset}} & \textbf{Dim} & \textbf{Prediction Length} & \textbf{Dataset Size} & \textbf{Frequency} & \textbf{Information}  \\ \hline
		ETTh1, ETTh2 & 7 & \{96, 192, 336, 720\} & \{8545, 2881, 2881\} & Hourly & Electricity \\ \hline
		ETTm1, ETTm2 & 7 & \{96, 192, 336, 720\} & \{34465, 11521, 11521\} & 15min & Electricity \\ \hline
		Exchange & 8 & \{96, 192, 336, 720\} & \{5120, 665, 1422\} & Daily & Economy \\ \hline
		Weather & 21 & \{96, 192, 336, 720\} & \{36792, 5271, 10540\} & 10min & Weather \\ \hline
		ECL & 321 & \{96, 192, 336, 720\} & \{18317, 2633, 5261\} & 10min & Electricity \\ \hline
		Traffic & 862 & \{96, 192, 336, 720\} & \{12185, 1757, 3509\} & Hourly & Transportation \\ \hline
		Solar-Energy & 137 & \{96, 192, 336, 720\} & \{36601, 5161, 10417\} & 10min & Energy \\ \hline
		PEMS03 & 358 & \{12, 24, 48, 96\} & \{15671, 5135, 5135\} & 5min & Transportation \\ \hline
		PEMS04 & 307 & \{12, 24, 48, 96\} & \{10172, 3375, 3375\} & 5min & Transportation \\ \hline
		PEMS07 & 883 & \{12, 24, 48, 96\} & \{16911, 5622, 5622\} & 5min & Transportation \\ \hline
		PEMS08 & 170 & \{12, 24, 48, 96\} & \{10690, 3548, 3548\} & 5min & Transportation \\ \hline
		BTC/USDT & 5 & \{12, 288, 864\} & \{12989, 2004, 4007\} & 5min & Economy \\ \hline
	\end{tabular}
	\label{tab:1}
\end{table*}

\begin{table*}[!ht]
	\centering
	\scriptsize
	\setlength{\tabcolsep}{3.9pt}       
	\caption{Comparison of multivariate time series forecasting results for 13 real datasets. Average long-term forecast results with a uniform lookback window \(\mathcal{L}\) = 96 for all datasets. All results are averaged over 4 different forecast lengths: \(\mathcal{F}\) = \{12, 24, 48, 96\} for the PEMS dataset and \(\mathcal{F}\) = \{96, 192, 336, 720\} for all other datasets. The best model is shown in \textbf{bold black}, and the second-best is \underline{underlined}. See Appendix \ref{apB} for complete results.}
	\begin{tabular}{l|cc|cc|cc|cc|cc|cc|cc|cc|cc|cc|cc}
		\toprule
		\textbf{Models} & \multicolumn{2}{c|}{\makecell{Time-TK\\(Ours)}} & \multicolumn{2}{c|}{\makecell{MMK\\(2025)}} & \multicolumn{2}{c|}{\makecell{TimeKAN\\(2025)}} & \multicolumn{2}{c|}{\makecell{CMoS\\(2025)}} & \multicolumn{2}{c|}{\makecell{MSGNet\\(2024)}} & \multicolumn{2}{c|}{\makecell{iTransformer\\(2024)}} & \multicolumn{2}{c|}{\makecell{TimeMixer\\(2024)}} & \multicolumn{2}{c|}{\makecell{PatchTST\\(2023)}} & \multicolumn{2}{c|}{\makecell{TimesNet\\(2023)}} & \multicolumn{2}{c|}{\makecell{DLinear\\(2023)}} & \multicolumn{2}{c}{\makecell{Crossformer\\(2023)}} \\
		\cmidrule(r){2-3} \cmidrule(r){4-5} \cmidrule(r){6-7} \cmidrule(r){8-9} \cmidrule(r){10-11} \cmidrule(r){12-13} \cmidrule(r){14-15} \cmidrule(r){16-17} \cmidrule(r){18-19} \cmidrule(r){20-21}\cmidrule(r){22-23}
		\textbf{Metric} & MSE & MAE & MSE & MAE & MSE & MAE & MSE & MAE & MSE & MAE & MSE & MAE & MSE & MAE & MSE & MAE & MSE & MAE & MSE & MAE & MSE & MAE \\
		\midrule
		\textbf{ETTh1} & \underline{0.432} & \textbf{0.430} & 0.432 & 0.436 & \textbf{0.425} & \underline{0.430} & 0.448 & 0.442 & 0.453 & 0.453 & 0.463 & 0.454 & 0.458 & 0.445 & 0.469 & 0.454 & 0.458 & 0.450 & 0.456 & 0.452 & 0.529 & 0.522 \\
		\textbf{ETTh2} & \textbf{0.372} & \textbf{0.397} & 0.390 & 0.417 & 0.390 & 0.408 & 0.392 & 0.410 & 0.413 & 0.427 & \underline{0.383} & \underline{0.407} & 0.384 & 0.407 & 0.389 & 0.411 & 0.414 & 0.427 & 0.559 & 0.515 & 0.942 & 0.684 \\
		\textbf{ETTm1} & \textbf{0.379} & \textbf{0.393} & \underline{0.384} & 0.397 & \textbf{0.379} & \underline{0.396} & 0.412 & 0.410 & 0.400 & 0.412 & 0.405 & 0.410 & 0.385 & 0.399 & 0.396 & 0.406 & 0.400 & 0.406 & 0.403 & 0.407 & 0.513 & 0.495 \\
		\textbf{ETTm2} & \textbf{0.276} & \textbf{0.321} & \underline{0.278} & 0.327 & 0.279 & \underline{0.324} & 0.288 & 0.330 & 0.289 & 0.330 & 0.290 & 0.335 & 0.280 & 0.325 & 0.291 & 0.336 & 0.291 & 0.333 & 0.350 & 0.401 & 0.757 & 0.611 \\
		\textbf{Electricity} & \textbf{0.174} & \textbf{0.265} & 0.201 & 0.286 & 0.197 & 0.286 & 0.204 & 0.284 & 0.194 & 0.301 & \underline{0.178} & \underline{0.270} & 0.182 
		& 0.272 & 0.211 & 0.301 & 0.193 & 0.295 & 0.212 & 0.300 & 0.244 & 0.334 \\
		\textbf{Exchange} & \textbf{0.353} & \textbf{0.397} & 0.375 & \underline{0.412} & 0.404 & 0.423 & 0.388 & 0.427 & 0.399 & 0.430 & 0.375 & 0.412 & 0.408 & 0.422 & 0.378 & 0.415 & 0.416 & 0.443 & \underline{0.354} & 0.414 & 0.471 & 0.478 \\
		\textbf{Solar-Energy} & \textbf{0.205} & \textbf{0.257} & 0.243 & 0.299 & 0.287 & 0.321 & 0.332 & 0.322 & 0.263 & 0.292 & \underline{0.233} & \underline{0.262} & 0.237 & 0.290 & 0.270 & 0.307 & 0.301 & 0.319 & 0.330 & 0.401 & 0.641 & 0.639 \\
		\textbf{Weather} & 0.256 & 0.278 & 0.246 & \underline{0.273} & \textbf{0.243} & \textbf{0.272} & 0.251 & 0.278 & 0.249 & 0.278 & 0.258 & 0.278 & \underline{0.245} & 0.276 & 0.259 & 0.281 & 0.259 & 0.287 & 0.265 & 0.317 & 0.259 & 0.315 \\
		\textbf{Traffic} & \textbf{0.425} & \textbf{0.278} & 0.541 & 0.335 & 0.590 & 0.374 & 0.617 & 0.366 & 0.660 & 0.382 & \underline{0.428} & \underline{0.282} & 0.485 & 0.298 & 0.555 & 0.362 & 0.620 & 0.336 & 0.625 & 0.383 & 0.550 & 0.304 \\
		\textbf{PEMS03} & \textbf{0.112} & \textbf{0.219} & 0.158 & 0.261 & 0.171 & 0.258 & 0.147 & 0.253 & 0.150 & 0.251 & \underline{0.113} & \underline{0.221} & 0.144 & 0.258 & 0.137 & 0.240 & 0.147 & 0.248 & 0.278 & 0.375 & 0.169 & 0.281 \\
		\textbf{PEMS04} & \textbf{0.109} & \textbf{0.218} & 0.152 & 0.279 & 0.148 & 0.259 & 0.124 & 0.249 & 0.122 & 0.239 & \underline{0.111} & \underline{0.221} & 0.161 & 0.272 & 0.145 & 0.249 & 0.129 & 0.241 & 0.295 & 0.388 & 0.209 & 0.314 \\
		\textbf{PEMS07} & \textbf{0.093} & \textbf{0.195} & 0.138 & 0.233 & 0.139 & 0.240 & 0.154 & 0.247 & 0.122 & 0.227 & \underline{0.101} & \underline{0.204} & 0.162 & 0.253 & 0.144 & 0.233 & 0.124 & 0.225 & 0.329 & 0.395 & 0.235 & 0.315 \\
		\textbf{PEMS08} & \textbf{0.145} & \textbf{0.224} & 0.214 & 0.268 & 0.213 & 0.291 & 0.176 & 0.255 & 0.205 & 0.285 & \underline{0.150} & \underline{0.226} & 0.206 & 0.296 & 0.200 & 0.275 & 0.193 & 0.271 & 0.379 & 0.416 & 0.268 & 0.307 \\
		\midrule
		\textbf{Count} & \multicolumn{2}{c|}{\textbf{23}} & \multicolumn{2}{c|}{{0}} & \multicolumn{2}{c|}{\underline{4}} & \multicolumn{2}{c|}{0} & \multicolumn{2}{c|}{0} & \multicolumn{2}{c|}{{0}} & \multicolumn{2}{c|}{{0}} & \multicolumn{2}{c|}{0} & \multicolumn{2}{c|}{0} & \multicolumn{2}{c|}{0} & \multicolumn{2}{c}{0} \\
		\bottomrule
	\end{tabular}
	
	\label{tab:comparison}
\end{table*}

\section{Experiments}
\subsection{Experimental Setup}	
\textbf{Datasets.} To validate the effectiveness of Time-TK, we conducted extensive experiments on 14 different datasets, as shown in the Table \ref{tab:1}, including four subsets of ETT (ETTh1, ETTh2, ETTm1, and ETTm2), Electricity, Exchanges, Solar-Energy, weather \citep{8}, and Traffic. For short-term forecasting, we used four subsets of PEMS (PEMS03, PEMS04, PEMS07, and PEMS08). Furthermore, we included 20,000 BTC/USDT data records with a 5-minute throughput. Accurate forecasts can significantly improve the effectiveness of remedial or preventive measures implemented by web applications, such as in intelligent traffic management and website transactions. See Appendix \ref{apA.1} for more detailed information.

\textbf{Setup.} All experiments are implemented in PyTorch. We use the mainstream MSE and MAE as our evaluation indicators. See Appendix \ref{apA.2} for more detailed information.

\textbf{Baselines. }We select 10 latest models, including MMK \citep{han2024kans}, TimeKAN \citep{huang2025timekan}, CMoS \citep{si2025cmos}, MSGNet \citep{cai2024msgnet}, iTransformer \citep{15}, TimeMixer \citep{13}, PatchTST \citep{16}, TimesNet \citep{28}, DLinear \citep{10} and Crossformer \citep{25} as our baselines.

\begin{table*}[ht]
	\centering
	\caption{Performance comparison on the BTC/USDT dataset. We predict the transaction throughput for the next hour, day, and three days with an input length of \(\mathcal{L}\) = 96.}
	\label{tab:model_performance_beautified}
	\renewcommand{\arraystretch}{0.6}
	\setlength{\tabcolsep}{7pt} 
	\begin{tabular}{c|c|c|c|c|c|c|c|c}
		\toprule
		\textbf{Setting} & \textbf{Metric} & \textbf{Time-TK} & \textbf{  MMK   } & \textbf{TimeKAN} & \textbf{CMoS} & \textbf{MSGNet} & \textbf{iTransformer} & \textbf{TimeMixer} \\
		
		\midrule
		\multirow{4}{*}{\textbf{BTC/USDT->1 hour}} & MAE & \textbf{0.103} & 0.112 & \underline{0.105} & 0.112 & 0.114 & 0.112 & 0.109 \\
		
		& RSE & \textbf{0.725} & 0.742 & \textbf{0.725} & 0.729 & 0.732 & 0.729 & \underline{0.727} \\
		& RMSE & \textbf{0.402} & 0.418 & \underline{0.407} & 0.411 & 0.415 & 0.411 & 0.409 \\
		& MAPE & \underline{1.459} & 1.509 & \textbf{1.358} & 1.480 & 1.520 & 1.480 & 1.527 \\
		\midrule
		\multirow{4}{*}{\textbf{BTC/USDT->1 day}} & MAE & \textbf{0.228} & 0.237 & 0.232 & 0.242 & 0.242 & 0.238 & \underline{0.230} \\
		& RSE & \textbf{0.904} & 0.93 & {0.913} & 0.945 & 0.925 & 0.922 & \underline{0.910} \\
		& RMSE & \textbf{0.471} & 0.493 & 0.484 & 0.498 & \underline{0.483} & 0.489 & \underline{0.483} \\
		& MAPE & 3.606 & 3.661 & \underline{3.568} & 3.726 & 3.740 & 3.710 & \textbf{3.560} \\
		\midrule
		\multirow{4}{*}{\textbf{BTC/USDT->3 day}} & MAE & \textbf{0.324} & 0.340 & 0.332 & 0.336 & 0.341 & \underline{0.330} & 0.340 \\
		& RSE & \textbf{1.020} & 1.163 & 1.143 & 1.159 & 1.210 & \underline{1.140} & 1.169 \\
		& RMSE & \textbf{0.531} & 0.544 & 0.534 & 0.548 & 0.551 & \underline{0.533} & 0.547 \\
		& MAPE & \textbf{8.512} & 8.886 & \underline{8.534} & 8.868 & 8.893 & 8.728 & 8.806 \\
		\bottomrule
	\end{tabular}
\end{table*}
\subsection{Main Results}

The comprehensive prediction results of Time-TK and 13 baseline models are shown in Table \ref{tab:comparison}. The best results are marked in bold and the second best results are marked in black underline. The lower the MSE and MAE, the more accurate the prediction results. Time-TK ranked first in 23 of the 26 experimental cases, demonstrating its excellent performance in both long and short time series prediction tasks. On the Weather dataset, TimeKAN \citep{huang2025timekan} achieved the best results. This may be because Weather data has multiple periodic features such as seasonality and daily periodicity \citep{we}, and is accompanied by strong non-stationarity. The frequency decomposition architecture adopted by TimeKAN can effectively model periodic signals of different frequencies, so it performs particularly well on multi-period datasets such as Weather.

It is worth noting that we also compare with existing models based on KAN architecture. Compared with MMK \citep{han2024kans}, our model Time-TK reduces MSE by 6.69\% and MAE by 7.90\% on average on 13 real-world datasets. Compared with TimeKAN, Time-TK reduces MSE by 7.4\% and MAE by 8.57\%, indicating that Time-TK is successful in introducing KAN network into time series modeling. In addition, compared with iTransformer \citep{15} based on overall temporal embedding and PatchTST \citep{16} based on temporal patch embedding, Time-TK reduces MSE by 6.41\%/10.84\% and MAE by 5.47\%/10.71\%, respectively.

Additionally, Table \ref{tab:model_performance_beautified} presents the detailed prediction results of Time-TK against seven baseline models on the BTC/USDT dataset. The results clearly demonstrate a significant and consistent performance advantage for Time-TK across all prediction horizons. Out of 12 experimental evaluations (4 metrics across 3 settings), Time-TK secured the top result 8 times and placed within the top two 10 times. This strongly indicates that the Time-TK architecture is highly effective for modeling complex time series data.

For overall performance, we train on 10 representative datasets with 3 random seeds over 4 forecasting horizons, as shown in Table \ref{tab:overall_performance}. Using TimeKAN as the baseline, a Wilcoxon test on the averaged MSE/MAE yields $p$-values of $1.86\times 10^{-2}$ and $5.86\times 10^{-3}$, both below 0.02, indicating that the overall improvement of Time-TK over TimeKAN is statistically significant at the 98\% confidence level (99\% for MAE).

Overall, these significant performance improvements are mainly due to the synergy between our proposed Multi-Offset Token Embedding and Multi-Offset Interaction mechanism, which enables the model to effectively capture complex and multi-scale dynamic patterns in the time dimension.
\begin{table}[!ht]
	\centering
 \scriptsize
\renewcommand{\arraystretch}{0.72}
	\setlength{\tabcolsep}{5pt}
	\caption{Ablation experiments of multi-offset embedding and multi-offset interaction of Time-TK.}
	\begin{tabular}{llcccccccc}
		\toprule
		\multicolumn{2}{c}{\multirow{2}{*}{{Strategy}}} & \multicolumn{6}{c}{{Time-TK}} & \multicolumn{2}{c}{\multirow{2}{*}{{iTransformer}}} \\
		\cmidrule(lr){3-8}
		\multicolumn{2}{c}{} & \multicolumn{2}{c}{{MOTI+MOTE}} & \multicolumn{2}{c}{{MOTI}} & \multicolumn{2}{c}{{MOTE}} & \multicolumn{2}{c}{} \\
		\multicolumn{2}{c}{{Metric}} & MSE & MAE & MSE & MAE & MSE & MAE & MSE & MAE \\
		\midrule
		\multirow{4}{*}{ETTh1} & 96  & \textbf{0.370} & \textbf{0.393} & 0.378 & 0.396 & 0.416 & 0.427 & 0.394 & 0.409 \\
		& 192 & \textbf{0.423} & \textbf{0.421} & 0.433 & 0.424 & 0.468 & 0.457 & 0.448 & 0.441 \\
		& 336 & \textbf{0.465} & \textbf{0.444} & 0.473 & 0.444 & 0.510 & 0.479 & 0.491 & 0.464 \\
		& 720 & \textbf{0.470} & \textbf{0.462} & 0.493 & 0.470 & 0.490 & 0.489 & 0.519 & 0.502 \\
		\midrule
		\multirow{4}{*}{ETTm1} & 96  & \textbf{0.315} & \textbf{0.354} & 0.326 & 0.362 & 0.347 & 0.382 & 0.336 & 0.370 \\
		& 192 & \textbf{0.356} & \textbf{0.378} & 0.367 & 0.382 & 0.389 & 0.400 & 0.381 & 0.395 \\
		& 336 & \textbf{0.393} & \textbf{0.402} & 0.405 & 0.403 & 0.417 & 0.422 & 0.417 & 0.418 \\
		& 720 & \textbf{0.453} & \textbf{0.439} & 0.468 & 0.443 & 0.470 & 0.454 & 0.487 & 0.456 \\
		\midrule
		\multirow{4}{*}{Exchange} & 96  & \textbf{0.083} & \textbf{0.202} & 0.084 & 0.203 & 0.092 & 0.214 & 0.088 & 0.209 \\
		& 192 & \textbf{0.168} & \textbf{0.292} & 0.178 & 0.299 & 0.189 & 0.312 & 0.183 & 0.308 \\
		& 336 & \textbf{0.322} & \textbf{0.411} & 0.332 & 0.416 & 0.356 & 0.432 & 0.336 & 0.418 \\
		& 720 & \textbf{0.838} & \textbf{0.684} & 0.890 & 0.706 & 1.149 & 0.784 & 0.893 & 0.714 \\
		\midrule
		\multirow{4}{*}{PEMS08} & 12  & \textbf{0.076} & \textbf{0.175} & 0.085 & 0.186 & 0.091 & 0.195 & 0.079 & 0.182 \\
		& 24 & \textbf{0.106} & \textbf{0.206} & 0.122 & 0.226 & 0.123 & 0.224 & 0.115 & 0.219 \\
		& 48 & \textbf{0.183} & \textbf{0.251} & 0.199 & 0.276 & 0.160 & 0.272 & 0.186 & 0.235 \\
		& 96 & \textbf{0.215} & \textbf{0.262} & 0.245 & 0.302 & 0.242 & 0.300 & 0.221 & 0.267 \\
		\bottomrule
	\end{tabular}
	
	\label{tab:Ablation}
\end{table}

\begin{table}[ht]
	\centering
	\renewcommand{\arraystretch}{0.62}
	\setlength{\tabcolsep}{2pt}
	\scriptsize
	\caption{MOTE can effectively improve the forecasting performance of models with different embedding strategies.}
	\begin{tabular}{cc|cc|cc||cc|cc||cc|cc}
		\toprule
		\multicolumn{2}{c|}{\textbf{Model}} &
		\multicolumn{4}{c||}{\makecell{\textbf{iTransformer}\\Inverted Embedding}} &
		\multicolumn{4}{c||}{\makecell{\textbf{PatchTST}\\Patch Embedding}} &
		\multicolumn{4}{c}{\makecell{\textbf{TimesNet}\\Mixed Embedding}} \\
		
		\midrule
		\multicolumn{2}{c|}{\textbf{Setup}} 
		& \multicolumn{2}{c|}{Original} & \multicolumn{2}{c||}{+MOTE}
		& \multicolumn{2}{c|}{Original} & \multicolumn{2}{c||}{+MOTE}
		& \multicolumn{2}{c|}{Original} & \multicolumn{2}{c}{+MOTE} \\
		
		\midrule
		\multicolumn{2}{c|}{\textbf{Metric}} 
		& MSE & MAE & MSE & MAE 
		& MSE & MAE & MSE & MAE 
		& MSE & MAE & MSE & MAE \\
		
		\midrule
		\multirow{5}{*}{\rotatebox{90}{ETTh1}} 
		& 96  & 0.394 & 0.409 & \textbf{0.389} & \textbf{0.405} & 0.414 & 0.419 & \textbf{0.403} & \textbf{0.416} & 0.384 & 0.402 & \textbf{0.379} & \textbf{0.398} \\
		& 192 & 0.448 & 0.441 & \textbf{0.443} & \textbf{0.440} & 0.460 & 0.445 & \textbf{0.449} & \textbf{0.439} & 0.436 & 0.429 & \textbf{0.431} & \textbf{0.426} \\
		& 336 & 0.491 & 0.464 & \textbf{0.487} & \textbf{0.461} & 0.501 & 0.466 & \textbf{0.488} & \textbf{0.456} & 0.491 & 0.469 & \textbf{0.490} & \textbf{0.467} \\
		& 720 & 0.519 & 0.502 & \textbf{0.511} & \textbf{0.492} & 0.500 & 0.488 & \textbf{0.487} & \textbf{0.477} & 0.521 & 0.500 & \textbf{0.513} & \textbf{0.497} \\
		& Avg & 0.463 & 0.454 & \textbf{0.458} & \textbf{0.450} & 0.469 & 0.454 & \textbf{0.457} & \textbf{0.447} & 0.458 & 0.450 & \textbf{0.453} & \textbf{0.447} \\
		
		\midrule
		\multirow{5}{*}{\rotatebox{90}{ETTh2}} 
		& 96  & 0.297 & 0.349 & \textbf{0.296} & \textbf{0.345} & 0.292 & 0.345 & \textbf{0.292} & \textbf{0.344} & 0.340 & 0.374 & \textbf{0.312} & \textbf{0.364} \\
		& 192 & 0.380 & 0.400 & \textbf{0.375} & \textbf{0.397} & 0.388 & 0.405 & \textbf{0.376} & \textbf{0.393} & 0.402 & 0.414 & \textbf{0.386} & \textbf{0.403} \\
		& 336 & 0.428 & 0.432 & \textbf{0.419} & \textbf{0.429} & 0.427 & 0.436 & \textbf{0.382} & \textbf{0.410} & 0.452 & 0.452 & \textbf{0.423} & \textbf{0.432} \\
		& 720 & 0.427 & 0.445 & \textbf{0.419} & \textbf{0.438} & 0.447 & 0.458 & \textbf{0.411} & \textbf{0.433} & 0.462 & 0.468 & \textbf{0.448} & \textbf{0.462} \\
		& Avg & 0.383 & 0.407 & \textbf{0.377} & \textbf{0.402} & 0.389 & 0.411 & \textbf{0.365} & \textbf{0.395} & 0.414 & 0.427 & \textbf{0.392} & \textbf{0.415} \\
		
		\midrule
		\multirow{5}{*}{\rotatebox{90}{Exchange}} 
		& 96  & 0.088 & 0.209 & \textbf{0.088} & \textbf{0.208} & 0.090 & 0.211 & \textbf{0.084} & \textbf{0.202} & 0.107 & 0.234 & \textbf{0.091} & \textbf{0.211} \\
		& 192 & 0.183 & 0.308 & \textbf{0.179} & \textbf{0.304} & 0.186 & 0.307 & \textbf{0.174} & \textbf{0.296} & 0.226 & 0.344 & \textbf{0.192} & \textbf{0.321} \\
		& 336 & 0.336 & 0.418 & \textbf{0.321} & \textbf{0.411} & 0.339 & 0.424 & \textbf{0.320} & \textbf{0.407} & 0.367 & 0.448 & \textbf{0.362} & \textbf{0.403} \\
		& 720 & 0.893 & 0.714 & \textbf{0.864} & \textbf{0.707} & 0.898 & 0.718 & \textbf{0.855} & \textbf{0.696} & 0.964 & 0.746 & \textbf{0.912} & \textbf{0.721} \\
		& Avg & 0.375 & 0.412 & \textbf{0.363} & \textbf{0.408} & 0.378 & 0.415 & \textbf{0.358} & \textbf{0.400} & 0.416 & 0.443 & \textbf{0.389} & \textbf{0.414} \\
		
		\midrule
		\multirow{5}{*}{\rotatebox{90}{Solar}} 
		& 96  & 0.203 & 0.238 & \textbf{0.198} & \textbf{0.235} & 0.234 & 0.286 & \textbf{0.231} & \textbf{0.278} & 0.250 & 0.292 & \textbf{0.237} & \textbf{0.284} \\
		& 192 & 0.233 & \textbf{0.261} & \textbf{0.226} & 0.269 & 0.267 & 0.310 & \textbf{0.257} & \textbf{0.296} & 0.296 & 0.318 & \textbf{0.288} & \textbf{0.311} \\
		& 336 & 0.248 & \textbf{0.273} & \textbf{0.236} & \textbf{0.273} & 0.290 & 0.315 & \textbf{0.283} & \textbf{0.308} & 0.319 & 0.330 & \textbf{0.292} & \textbf{0.317} \\
		& 720 & 0.249 & \textbf{0.276} & \textbf{0.243} & 0.283 & 0.289 & 0.317 & \textbf{0.286} & \textbf{0.313} & 0.338 & 0.337 & \textbf{0.301} & \textbf{0.319} \\
		& Avg & 0.233 & \textbf{0.262} & \textbf{0.226} & {0.265} & 0.270 & 0.307 & \textbf{0.264} & \textbf{0.299} & 0.301 & 0.319 & \textbf{0.280} & \textbf{0.308} \\
		\bottomrule
	\end{tabular}
	
	\label{tab:qianyi}
\end{table}

\subsection{Model Analysis}
\subsubsection{Ablation Study on the Design of MOTE}
As shown in Table \ref{tab:Ablation}, we further evaluated the independent contribution of each component to the model performance. By comparing the prediction results after removing the two key components of Time-TK, Multi-Offset Token Embedding (MOTE) and Multi-Offset Temporal Interaction (MOTI), we found that these two components have a significant positive effect on improving prediction efficiency.

\begin{figure*}[!ht]    
	\centering
	\begin{minipage}[b]{0.99\textwidth} 
		\includegraphics[width=\textwidth]{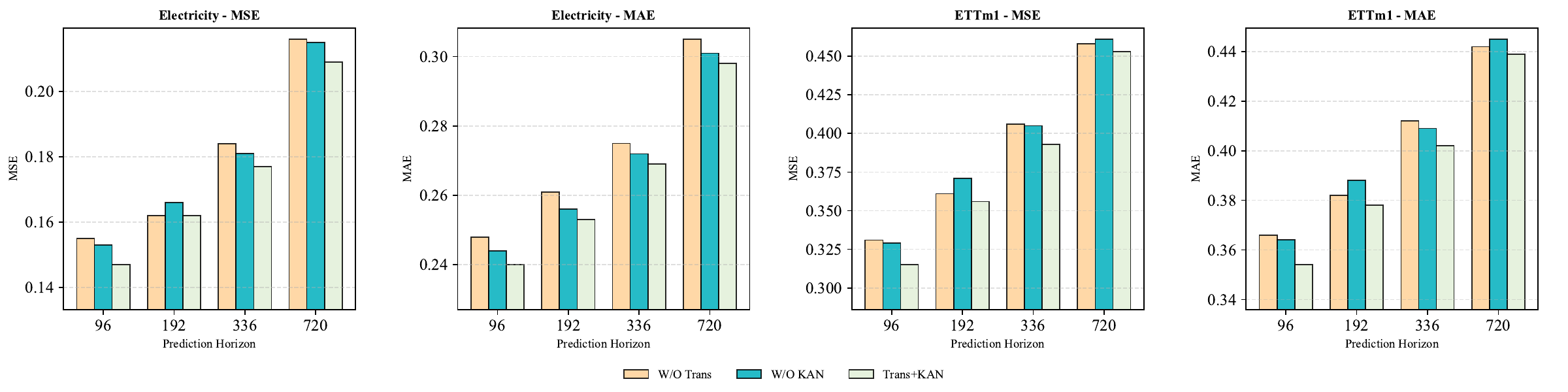}
		\centering
	\end{minipage}
	\caption{Ablation study comparing Time-TK with its architectural variants on the Electricity and ETTm1 datasets across multiple prediction horizons.}
	\label{fig:KT}
\end{figure*}
\begin{table}[ht]
	\centering
	\scriptsize
	\setlength{\tabcolsep}{1.6pt}
		\caption{Ablation experiments of MI-KAN of Time-TK.}
	\begin{tabular}{l|cc|cc|cc|cc|cc|cc}
		\toprule
		\makecell[l]{\textbf{Metric}} & 
		\multicolumn{2}{c|}{\textbf{ETTh1}} & 
		\multicolumn{2}{c|}{\textbf{ETTh2}} & 
		\multicolumn{2}{c|}{\textbf{ETTm1}} & 
		\multicolumn{2}{c|}{\textbf{ETTm2}} & 
		\multicolumn{2}{c|}{\textbf{Solar-Energy}} & 
		\multicolumn{2}{c}{\textbf{Electricity}} \\
		\textbf{Datasets} & MSE & MAE & MSE & MAE & MSE & MAE & MSE & MAE & MSE & MAE & MSE & MAE \\
		\midrule
		MLP & 0.379 & 0.396 & 0.298 & 0.345 & 0.320 & 0.357 & 0.176 & 0.260 & 0.217 & 0.289 & 0.155 & 0.247 \\
		Conv1d & 0.375 & 0.394 & 0.300 & 0.346 & 0.318 & 0.357 & 0.176 & 0.257 & 0.214 & 0.287 & 0.161 & 0.252 \\
		KAN & 0.376 & 0.396 & 0.295 & 0.343 & 0.319 & 0.356 & 0.175 & 0.255 & 0.210 & 0.281 & 0.151 & 0.244 \\
		\textbf{RBF} & \textbf{0.370} & \textbf{0.393} & \textbf{0.293} & \textbf{0.340} & \textbf{0.315} & \textbf{0.354} & \textbf{0.173} & \textbf{0.253} & \textbf{0.187} & \textbf{0.234} & \textbf{0.147} & \textbf{0.240} \\
		\bottomrule
	\end{tabular}

	\label{tab:kan}
\end{table}
\subsubsection{Ablation Study on the Design of MI-KAN}
In this section, we design several variants to investigate the effectiveness of MI-KAN:
\textcircled{1} \textbf{MLP}, where each MI-KAN is replaced with a multilayer perceptron;
\textcircled{2} \textbf{Conv1d}, where each MI-KAN is replaced with a 1D convolutional layer;
\textcircled{3} \textbf{KAN}, which uses a B-spline-based KAN structure \citep{han2024kans};
\textcircled{4} \textbf{RBF}, where radial basis functions (RBFs) \citep{li2024kolmogorov, bresson2024kagnns} are used as the activation module in our MI-KAN. As shown in Table \ref{tab:kan}, MI-KAN achieves the best results. Notably, both MI-KAN and the B-spline-based KAN outperform MLP, indicating that KAN has stronger representational capacity than MLP. Moreover, MI-KAN outperforms the B-spline-based KAN, further validating the effectiveness of adopting RBFs.

\begin{table}[ht]
	\centering

	\setlength{\tabcolsep}{3.5pt}
	\caption{Prediction performance under different input lengths on the ETTm1 dataset. The input length is selected as \(\mathcal{L}\)=\{48,96,144,192,288,384,480\}, and the fixed prediction length is \(\mathcal{F}\)=96. MOTE can effectively enhance the learning of historical information.}
	\begin{tabular}{c|cc|cc|cc|cc}
		\toprule
		\textbf{Model} & \multicolumn{2}{c|}{\textbf{PatchTST}} & \multicolumn{2}{c|}{\textbf{+MOTE}} & \multicolumn{2}{c|}{\textbf{iTrans}} & \multicolumn{2}{c}{\textbf{+MOTE}} \\
		\cmidrule(r){2-3} \cmidrule(r){4-5} \cmidrule(r){6-7} \cmidrule(r){8-9}
		\textbf{Metric}& MSE & MAE & MSE & MAE & MSE & MAE & MSE & MAE \\
		\midrule
		48  & 0.502 & 0.437 & 0.504 & 0.440 & 0.458 & 0.424 & 0.450 & 0.420 \\
		96  & 0.329 & 0.365 & 0.340 & 0.368 & 0.336 & 0.370 & 0.342 & 0.375 \\
		144 & 0.324 & 0.359 & 0.316 & 0.356 & 0.318 & 0.363 & 0.311 & 0.361 \\
		192 & 0.307 & 0.347 & 0.304 & 0.348 & 0.316 & 0.363 & 0.306 & 0.358 \\
		288 & 0.296 & 0.344 & 0.292 & 0.343 & 0.303 & 0.357 & 0.305 & 0.358 \\
		384 & 0.298 & 0.345 & 0.291 & 0.343 & 0.310 & 0.364 & 0.305 & 0.358 \\
		480 & 0.297 & 0.346 & 0.291 & 0.342 & 0.314 & 0.366 & 0.304 & 0.357 \\
		\bottomrule
	\end{tabular}
	
	\label{tab:mote_scaling}
\end{table}

\begin{figure}[!ht]    
	\centering
	\begin{minipage}[b]{0.48\textwidth} 
		\includegraphics[width=\textwidth]{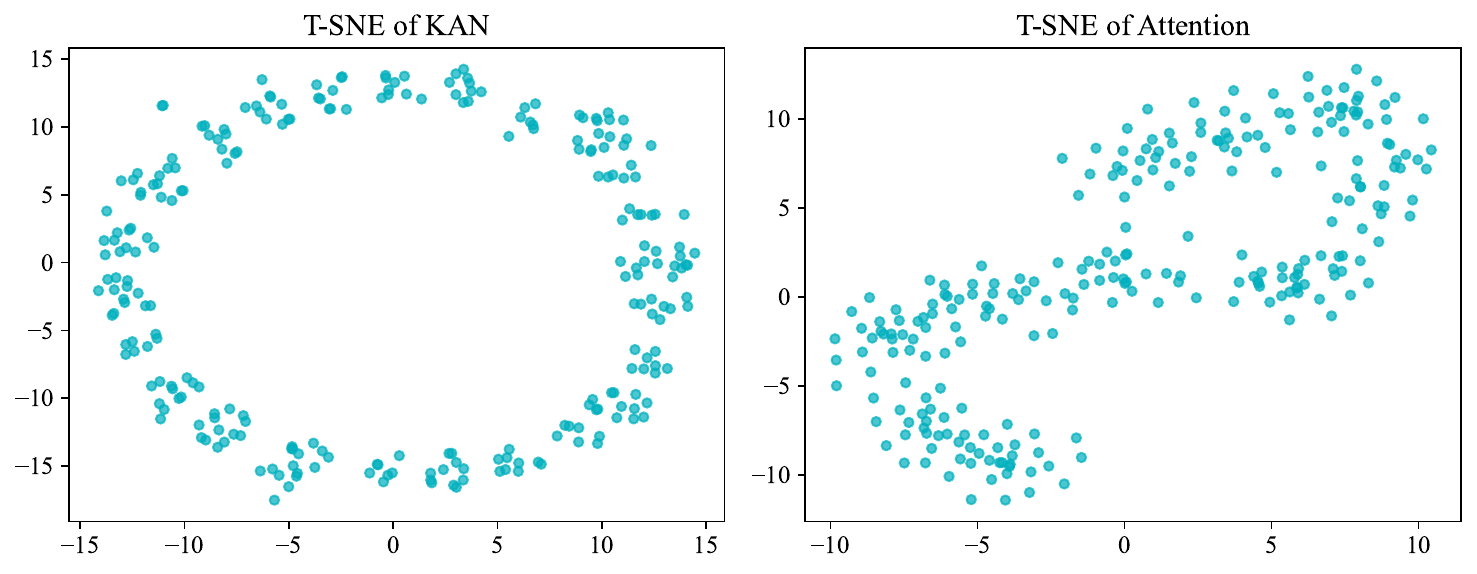}
		\centering
	\end{minipage}
	\caption{t-SNE visualization after KAN and Transformer.}
	\label{fig:tkt}
\end{figure}

\subsection{Effectiveness of combining KAN and Transformer}
To validate the synergistic effectiveness of the two core components in Time-TK, Transformer and KAN, we conducted a comprehensive ablation study, with the results shown in Figure~\ref{fig:KT}. In the experiment, we compared the full model (Trans+KAN) with two variants: one with the Transformer attention mechanism removed (W/O Trans), and another with the KAN module removed (W/O KAN). The results clearly indicate that our full model achieved the optimal results across all prediction horizons. This finding strongly demonstrates that removing either key component leads to a significant degradation in model performance, thus confirming the indispensability of both modules.  We further investigate this synergy by visualizing intermediate feature representations with t-SNE, as shown in Fig. tkt. As the first module in our architecture, KAN (Figure~\ref{fig:tkt}, left) acts as a nonlinear feature extractor that maps the raw input sequence onto a ring-shaped manifold with clear periodic structure, revealing the underlying nonlinear patterns before temporal dependencies are modeled. The subsequent Transformer attention (Figure~\ref{fig:tkt}, right) then performs weighted aggregation over multiple time steps on this manifold to capture long-range dependencies, which leads to a more “mixed” cloud in the t-SNE projection rather than strictly separated clusters. Overall, this visualization suggests that MOTE and KAN together organize the raw series into structured continuous representations and support multi-scale temporal integration on top of them.


\subsection{Performance promotion}
In addition to the ablation study on Multi-Offset Token Embedding (MOTE), we further evaluate its generalizability and transferability across different models. Specifically, we integrate MOTE into three representative models with different embedding strategies to verify whether it can enhance performance:

i) For iTransformer \citep{15} with holistic embedding, we apply MOTE for embedding and interaction before the attention module to enhance intra-sequence modeling capability;
ii) For PatchTST \citep{16} with patch embedding, we apply multi-offset sequences to its patch attention, enabling finer-grained modeling of the sequence structure;
iii) For TimesNet \citep{28}  with channel-mixing architecture, we introduce MOTE before the convolution operations, allowing the model to better capture complex periodic patterns.

As shown in Table \ref{tab:qianyi}, integrating the multi-offset embedding into all three architectures consistently improves forecasting performance, demonstrating that MOTE can be widely applied to various prediction models and that our proposed embedding mechanism exhibits strong scalability.

\begin{figure}[!ht]    
	\centering
	\begin{minipage}[b]{0.235\textwidth} 
		\includegraphics[width=\textwidth]{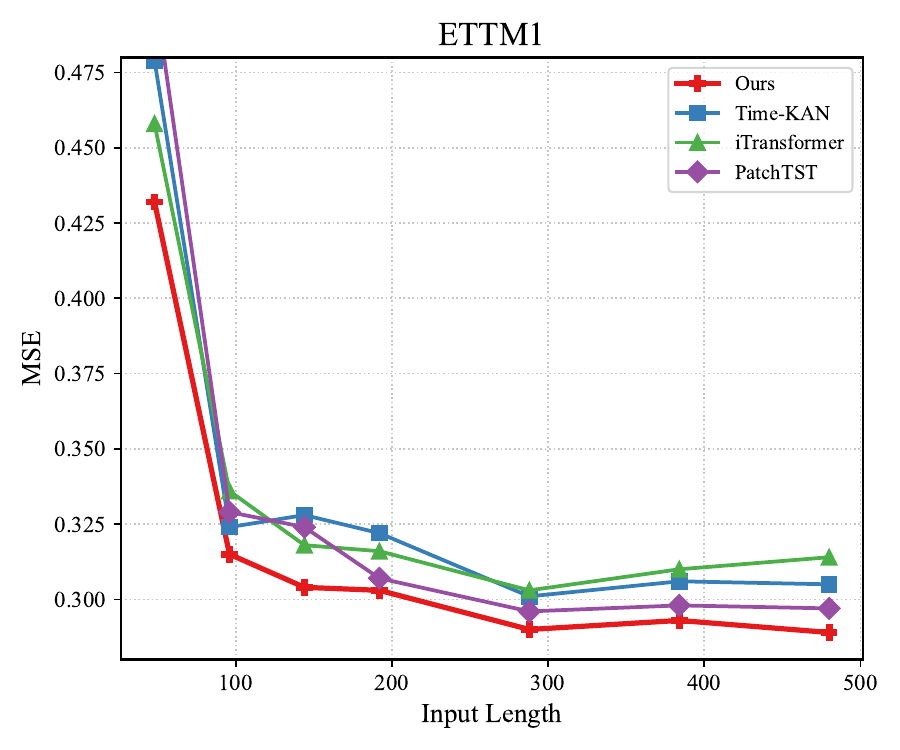}
		\centering
	\end{minipage}
	\hfill
	\begin{minipage}[b]{0.235\textwidth} 
		\includegraphics[width=\textwidth]{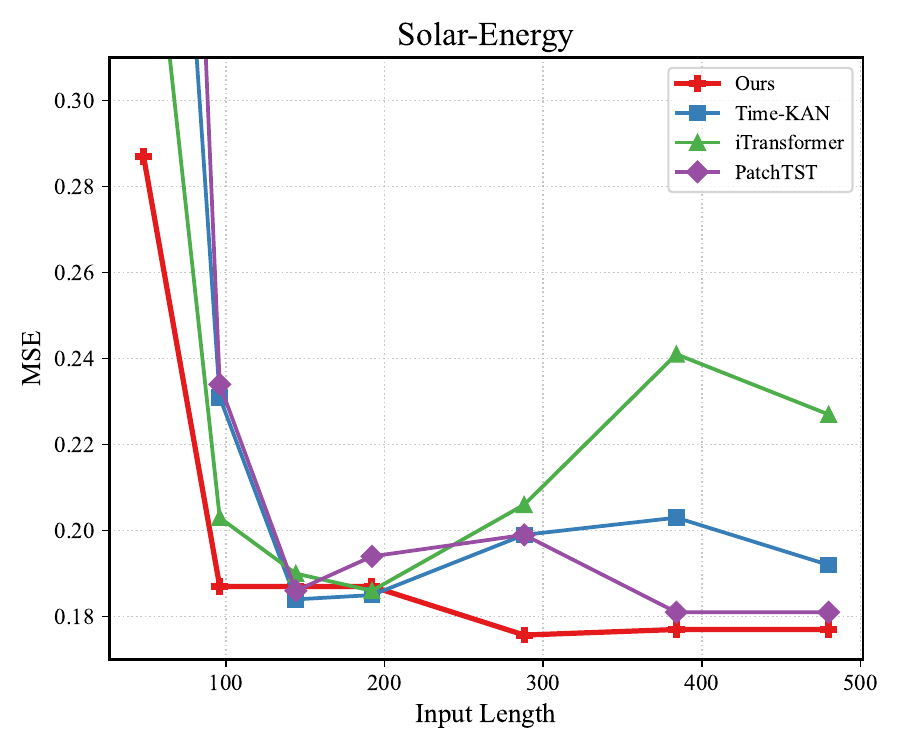}
		\centering
	\end{minipage}
	\caption{The prediction performance under different input lengths on two datasets. The input lengths are selected as \(\mathcal{L}\)=\{48,96,144,192,288,384,480\}, with a fixed prediction length of \(\mathcal{F}\)=96}
	\label{fig:input}
\end{figure}

\subsection{Increasing lookback Window}
Theoretically, based on statistical methods \citep{box1968some}, rich historical information helps models capture long-term dependencies in time series. A well-designed forecasting model should be able to effectively leverage longer historical sequences to improve predictive performance. As shown in Figure \ref{fig:input}, with the increase of input length, the iTransformer \citep{15} using full-sequence token embedding shows a significant drop in prediction accuracy, indicating that directly embedding the entire sequence may overlook finer-grained local information within the sequence, thus limiting modeling capacity for long inputs. In contrast, models using patch token embedding, such as TimeKAN \citep{huang2025timekan}, and PatchTST \citep{16}, exhibit more stable performance as the input length increases, suggesting that the patching mechanism helps mitigate performance degradation from long inputs. However, the increased number of patches also leads to higher memory costs, which limits scalability in long sequence modeling. Notably, our Time-TK benefits from the MOTE embedding strategy, where we embed multi-offset sub-sequences independently, enabling the model to adapt to longer lookback windows while maintaining low computational cost.

Previous studies show that the predictive performance of Transformer models does not necessarily improve with longer lookback lengths \citep{10}. Therefore, we introduce MOTE into two attention-based models, PatchTST and iTransformer. As shown in Table \ref{tab:mote_scaling}, the original models exhibit a general performance drop as input length increases, while after incorporating MOTE, both models surprisingly benefit from the extended historical window more effectively.

\subsection{Computational Cost}
\begin{figure}[!ht]    
	\centering
	\begin{minipage}[b]{0.45\textwidth} 
		\includegraphics[width=\textwidth]{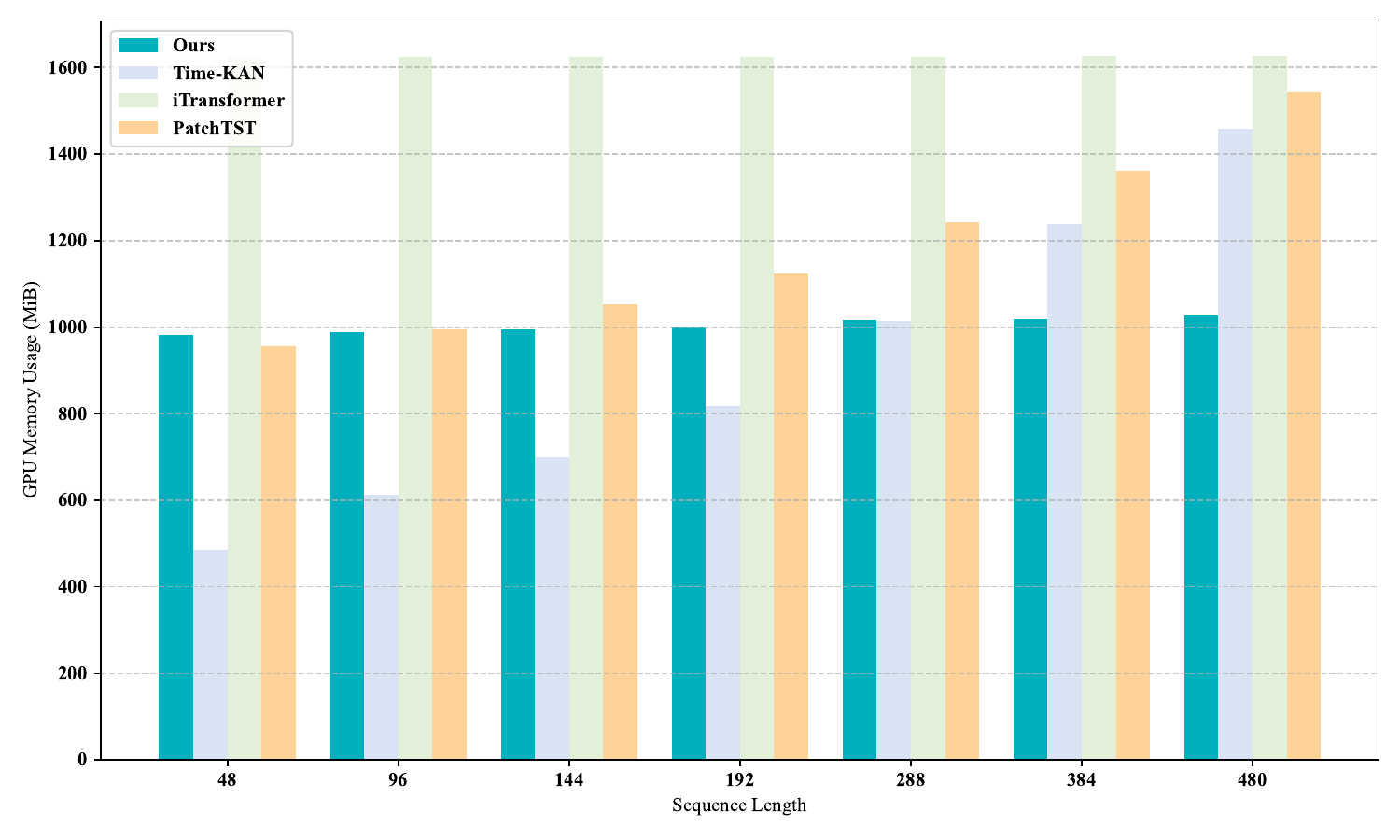}
		\centering
	\end{minipage}
	\caption{Comparison of GPU memory usage of different models at different input sequence lengths.}
	\label{fig:eff}
\end{figure}

To evaluate the computational cost of different models, we compare their GPU memory usage under varying input sequence lengths, as shown in Figure~\ref{fig:eff}. iTransformer consistently maintains high GPU memory consumption across all input lengths, with minimal variation, primarily due to its full-sequence embedding strategy, which makes its computational complexity less sensitive to input length. In contrast, the memory usage of PatchTST and TimeKAN increases significantly with longer sequences, owing to their patch-based embedding strategies, where the growing number of patches leads to higher memory overhead. Notably, our proposed Time-TK demonstrates excellent memory efficiency across different input lengths while still achieving superior predictive performance (see Figure \ref{fig:input}). This indicates that the designed Multi-Offset Token Embedding (MOTE) strategy can effectively utilize long historical information without introducing substantial computational burden, making Time-TK an efficient forecasting framework well-suited for long-sequence modeling.

\section{Conclusion}
This paper proposes a novel time series forecasting framework, Time-TK, which combines a Multi-Offset Token Embedding (MOTE) strategy with Multi-Offset Temporal Interaction. Unlike existing mainstream embedding methods, MOTE models the input sequence at multiple offset positions, enabling more efficient capture of both local and global temporal dynamics. It enhances the utilization of long historical information without significantly increasing memory overhead. Specifically, we first apply multi-offset embedding and then perform multi-offset temporal interaction to learn temporal dependencies across different time spans. Extensive experiments on 13 public datasets demonstrate that Time-TK outperforms existing state-of-the-art methods in both prediction accuracy and generalization capability. Notably, the MOTE embedding not only improves our model’s performance but also shows consistent gains when integrated into other architectures with different embedding schemes, further validating its generality and effectiveness. Time-TK offers new insights and directions for designing efficient and scalable time series forecasting models.
\section*{Acknowledgements}
This work was supported in part by the following: the Joint Fund of the National Natural Science Foundation of China under Grant Nos. U24A20219, U24A20328, U22A2033, the National Natural Science Foundation of China under Grant No. 62272281, the Special Funds for Taishan Scholars Project under Grant No. tsqn202306274, and the Youth Innovation Technology Project of Higher School in Shandong Province under Grant No. 2023KJ212.
\bibliographystyle{ACM-Reference-Format}
\bibliography{main}

\appendix
\section{Experimental Setting}

\subsection{DATASET DESCRIPTIONS}\label{apA.1}
To demonstrate the application effectiveness of Time-TK across various domains, we evaluated it on 14 different datasets. The descriptions of these datasets are as follows:

\begin{itemize}
	\item \textbf{ETT}\footnote{\url{https://github.com/zhouhaoyi/ETDataset}}\citep{24}: contains data with two different time granularities: one at the hourly level (ETTh) and the other at the 15-minute level (ETTm). This dataset records the temperature and load characteristics of seven oil transformers from July 2016 to July 2018.The Traffic dataset describes road occupancy, collected by sensors on the Francisco Freeway between 2015 and 2016, with hourly recordings.
	
	\item \textbf{Exchange}\footnote{\url{https://github.com/laiguokun/multivariate-time-series-data}}\citep{8}: A panel dataset of daily exchange rates from eight countries, spanning from 1990 to 2016.
	
	\item \textbf{Weather}\footnote{\url{https://www.bgc-jena.mpg.de/wetter/}}\citep{8}: Records 21 weather indicators, such as air temperature and humidity, with data collected every 10 minutes in 2020, from locations in Germany.
	
	\item \textbf{Electricity}\footnote{\url{https://archive.ics.uci.edu/ml/datasets/ElectricityLoadDiagrams20112014}}\citep{8}: Includes hourly electricity consumption data from 321 customers between 2012 and 2014.

	\item \textbf{Solar Energy}\footnote{\url{https://www.nrel.gov/grid/solar-power-data.html}}: Contains solar power generation data from 137 photovoltaic plants in 2006, sampled every 10 minutes.

	\item \textbf{PEMS}\footnote{\url{https://pems.dot.ca.gov/}}\citep{liu2022scinet}: Consists of California public transportation network data, collected in 5-minute windows. We use the same four public subsets (PEMS03, PEMS04, PEMS07, PEMS08) as those used in SCINet \citep{liu2022scinet}.
	
	\item \textbf{Traffic}\footnote{\url{https://pems.dot.ca.gov/}}\citep{8}: Collects hourly road occupancy data from 862 sensors on the San Francisco Freeway between January 2015 and December 2016.
	
	\item \textbf{BTC/USDT}\footnote{\url{https://www.kaggle.com/datasets/shivaverse/btcusdt-5-minute-ohlc-volume-data-2017-2025}}: The latest 20,000 BTC/USDT data records, including the opening price, highest price, lowest price, closing price, and actual trading volume of each record.

\end{itemize}

The main details are provided in Table \ref{tab:1}.

\begin{table*}[!h]   
	\caption{All experiments fix the lookback length \(\mathcal{L}\) = 96. For PEMS, the prediction length is \(\mathcal{F}\) \( \in \)\{12, 24, 48, 96\}; for other datasets, the prediction length is \(\mathcal{F}\) \( \in \) \{96, 192, 336, 720\}. The best result is \textbf{bold}, the second best result is \underline{underlined}.}
	\centering
	\renewcommand{\arraystretch}{0.57}
	\scriptsize 
	\setlength{\tabcolsep}{3.5pt} 
	\begin{tabular}{cccccccccccccccccccccccc}
		\toprule[1.2pt]
		\multirow{1}{*}{\textbf{Models}} & \multirow{2}{*}{\textbf{}} & \multicolumn{2}{c}{\textbf{Ours}} & \multicolumn{2}{c}{\textbf{MMK}} & \multicolumn{2}{c}{\textbf{TimeKAN}}& \multicolumn{2}{c}{\textbf{CMoS}} & \multicolumn{2}{c}{\textbf{MSGNet}} & \multicolumn{2}{c}{\textbf{iTrans}} & \multicolumn{2}{c}{\textbf{TimeMixer}}& \multicolumn{2}{c}{\textbf{PatchTST}} & \multicolumn{2}{c}{\textbf{TimesNet}} & \multicolumn{2}{c}{\textbf{DLinear}}& \multicolumn{2}{c}{\textbf{Crossformer}}  \\

		\cmidrule(r){1-2}	\cmidrule(r){3-4} \cmidrule(r){5-6} \cmidrule(r){7-8} \cmidrule(r){9-10} \cmidrule(r){11-12} \cmidrule(r){13-14} \cmidrule(r){15-16} \cmidrule(r){17-18} \cmidrule(r){19-20}\cmidrule(r){21-22}\cmidrule(r){23-24}
		\multirow{1}{*}{\textbf{Metrics}} 
		&{\textbf{}} & \textbf{MSE} & \textbf{MAE} & \textbf{MSE} & \textbf{MAE} & \textbf{MSE} & \textbf{MAE} & \textbf{MSE} & \textbf{MAE} & \textbf{MSE} & \textbf{MAE} & \textbf{MSE} & \textbf{MAE} & \textbf{MSE} & \textbf{MAE} & \textbf{MSE} & \textbf{MAE} & \textbf{MSE} & \textbf{MAE} & \textbf{MSE} & \textbf{MAE}& \textbf{MSE} & \textbf{MAE}\\
		\midrule
		\multirow{5}{*}{\textbf{ETTh1}} 
		& 96 & \textbf{0.370} & \textbf{0.393} & 0.374 & \underline{0.397} & \underline{0.373} & 0.397 & 0.396 & 0.403 & 0.390 & 0.411 & 0.394 & 0.409 & 0.381 & 0.401 & 0.414 & 0.419 & 0.384 & 0.402 & 0.386 & 0.400 & 0.423 & 0.448 \\
		& 192 & 0.423 & \textbf{0.421} & \underline{0.419} & 0.429 & \textbf{0.414} & \underline{0.421} & 0.432 & 0.428 & 0.443 & 0.442 & 0.448 & 0.441 & 
		0.440 & 0.433 & 0.460 & 0.445 & 0.436 & 0.429 & 0.437 & 0.432 & 0.471 & 0.474 \\
		& 336 & 0.465 & \underline{0.444} & \underline{0.461} & 0.450 & \textbf{0.451} & \textbf{0.442} & 0.481 & 0.454 & 0.482 & 0.469 & 0.491 & 0.464 & 
		0.501 & 0.462 & 0.501 & 0.466 & 0.491 & 0.469 & 0.481 & 0.459 & 0.570 & 0.546 \\
		& 720 & \underline{0.470} & \underline{0.462} & 0.474 & 0.467 & \textbf{0.460} & \textbf{0.460} & 0.482 & 0.482 & 0.496 & 0.488 & 0.519 & 0.502 & 
		0.501 & 0.482 & 0.500 & 0.488 & 0.521 & 0.500 & 0.519 & 0.516 & 0.653 & 0.621 \\
		& avg & \underline{0.432} & \textbf{0.430} & 0.432 & 0.436 & \textbf{0.425} & \underline{0.430} & 0.448 & 0.442 & 0.453 & 0.453 & 0.463 & 0.454 & 
		0.458 & 0.445 & 0.469 & 0.454 & 0.458 & 0.450 & 0.456 & 0.452 & 0.529 & 0.522 \\
		\midrule
		\multirow{5}{*}{\textbf{ETTh2}} 
		& 96 & 0.293 & \textbf{0.340} & 0.301 & 0.353 & 0.293 & \underline{0.341} & 0.309 & 0.351 & 0.329 & 0.371 & 0.297 & 0.349 & \textbf{0.292} & 0.343 & \underline{0.292} & 0.345 & 0.340 & 0.374 & 0.333 & 0.387 & 0.745 & 0.584 \\
		& 192 & \textbf{0.368} & \textbf{0.388} & 0.379 & 0.405 & \underline{0.377} & \underline{0.391} & 0.396 & 0.404 & 0.402 & 0.414 & 0.380 & 0.400  
		&0.378	&0.397 & 0.388 & 0.405 & 0.402 & 0.414& 0.477& 0.476 & 0.877    & 0.656  \\
		& 336 & \textbf{0.410} & \textbf{0.423} & 0.432 & 0.446 & \underline{0.423} & 0.435 & 0.431 & 0.437 & 0.440 & 0.445 & 0.428 & \underline{0.432} & 
		0.432 & 0.434 & 0.427 & 0.436 & 0.452 & 0.452 & 0.594 & 0.541 & 1.043 & 0.731 \\
		& 720 & \textbf{0.418} & \textbf{0.438} & 0.446 & 0.463 & 0.467 & 0.465 & 0.431 & 0.446 & 0.480 & 0.477 & \underline{0.427} & \underline{0.445} & 
		0.454 & 0.458 & 0.447 & 0.458 & 0.462 & 0.468 & 0.831 & 0.657 & 1.104 & 0.763 \\
		& avg & \textbf{0.372} & \textbf{0.397} & 0.390 & 0.417 & 0.390 & 0.408 & 0.392 & 0.410 & 0.413 & 0.427 & \underline{0.383} & \underline{0.407} & 
		0.384 & 0.407 & 0.389 & 0.411 & 0.414 & 0.427 & 0.559 & 0.515 & 0.942 & 0.684 \\
		\midrule
		\multirow{5}{*}{\textbf{ETTm1}} 
		& 96 & \textbf{0.315} & \textbf{0.354} & 0.320 & \underline{0.358} & 0.324 & 0.361 & 0.354 & 0.381 & \underline{0.319} & 0.366 & 0.336 & 0.370 & 0.328 & 0.363 & 0.329 & 0.365 & 0.338 & 0.375 & 0.345 & 0.372 & 0.404 & 0.426 \\
		& 192 & \textbf{0.356} & \textbf{0.378} & 0.364 & \underline{0.383} & \underline{0.357} & 0.383 & 0.390 & 0.396 & 0.377 & 0.397 & 0.381 & 0.395 & 
		0.364 & 0.384 & 0.380 & 0.394 & 0.374 & 0.387 & 0.380 & 0.389 & 0.450 & 0.451 \\
		& 336 & 0.393 & \textbf{0.402} & 0.395 & 0.405 & \textbf{0.386} & \underline{0.404} & 0.423 & 0.418 & 0.417 & 0.422 & 0.417 & 0.418 & \underline{0.390} & 0.404 & 0.400 & 0.410 & 0.410 & 0.411 & 0.413 & 0.413 & 0.532 & 0.515 \\
		& 720 & \underline{0.453} & \underline{0.439} & 0.457 & 0.440 & \textbf{0.447} & \textbf{0.437} & 0.481 & 0.445 & 0.487 & 0.463 & 0.487 & 0.456 & 
		0.458 & 0.445 & 0.475 & 0.453 & 0.478 & 0.450 & 0.474 & 0.453 & 0.666 & 0.589 \\
		& avg & \textbf{0.379} & \textbf{0.393} & 0.384 & 0.397 & \underline{0.379} & \underline{0.396} & 0.412 & 0.410 & 0.400 & 0.412 & 0.405 & 0.410 & 
		0.385 & 0.399 & 0.396 & 0.406 & 0.400 & 0.406 & 0.403 & 0.407 & 0.513 & 0.495 \\
		
		\midrule
		\multirow{5}{*}{\textbf{ETTm2}} 
		& 96 & \textbf{0.173} & \textbf{0.253} & 0.176 & 0.261 & \underline{0.174} & \underline{0.255} & 0.186 & 0.270 & 0.182 & 0.266 & 0.185 & 0.271 & 0.178 & 0.259 & 0.193 & 0.280 & 0.187 & 0.267 & 0.193 & 0.292 & 0.287 & 0.366 \\
		& 192 & \textbf{0.238} & \textbf{0.298} & 0.240 & 0.302 & \underline{0.239} & \underline{0.299} & 0.248 & 0.307 & 0.248 & 0.306 & 0.251 & 0.312 & 
		0.242 & 0.303 & 0.246 & 0.307 & 0.249 & 0.309 & 0.284 & 0.362 & 0.414 & 0.492 \\
		& 336 & \textbf{0.298} & \textbf{0.339} & \underline{0.299} & \underline{0.342} & 0.305 & 0.343 & 0.308 & 0.344 & 0.312 & 0.346 & 0.314 & 0.350 & 
		0.304 & 0.342 & 0.314 & 0.351 & 0.321 & 0.351 & 0.369 & 0.427 & 0.597 & 0.542 \\
		& 720 & \textbf{0.395} & \textbf{0.395} & 0.397 & 0.401 & 0.399 & 0.400 & 0.409 & 0.400 & 0.414 & 0.404 & 0.411 & 0.405 & \underline{0.395} & \underline{0.397} & 0.410 & 0.405 & 0.408 & 0.403 & 0.554 & 0.522 & 1.730 & 1.042 \\
		& avg & \textbf{0.276} & \textbf{0.321} & \underline{0.278} & 0.327 & 0.279 & \underline{0.324} & 0.288 & 0.330 & 0.289 & 0.330 & 0.290 & 0.335 & 
		0.280 & 0.325 & 0.291 & 0.336 & 0.291 & 0.333 & 0.350 & 0.401 & 0.757 & 0.611 \\
		
		\midrule
		\multirow{5}{*}{\textbf{Electricity}} 
		& 96 & \textbf{0.147} & \textbf{0.240} & 0.166 & 0.256 & 0.174 & 0.266 & 0.179 & 0.262 & 0.165 & 0.274 & \underline{0.148} & \underline{0.240} & 0.153 & 0.244 & 0.188 & 0.280 & 0.168 & 0.272 & 0.197 & 0.282 & 0.219 & 0.314 \\
		& 192 & \textbf{0.162} & \textbf{0.253} & 0.187 & 0.274 & 0.182 & 0.273 & 0.186 & 0.269 & 0.185 & 0.292 & \underline{0.162} & \underline{0.253} & 
		0.166 & 0.256 & 0.193 & 0.285 & 0.184 & 0.289 & 0.196 & 0.285 & 0.231 & 0.322 \\
		& 336 & \textbf{0.177} & \textbf{0.269} & 0.204 & 0.290 & 0.197 & 0.286 & 0.202 & 0.285 & 0.197 & 0.304 & \underline{0.178} & \underline{0.269} & 
		0.184 & 0.275 & 0.211 & 0.302 & 0.198 & 0.300 & 0.209 & 0.301 & 0.246 & 0.337 \\
		& 720 & \textbf{0.209} & \textbf{0.298} & 0.247 & 0.323 & 0.236 & 0.320 & 0.247 & 0.321 & 0.231 & 0.332 & 0.225 & 0.317 & 0.226 & \underline{0.313} & 0.253 & 0.335 & \underline{0.220} & 0.320 & 0.245 & 0.333 & 0.280 & 0.363 \\
		& avg & \textbf{0.174} & \textbf{0.265} & 0.201 & 0.286 & 0.197 & 0.286 & 0.204 & 0.284 & 0.194 & 0.301 & \underline{0.178} & \underline{0.270} & 
		0.182 & 0.272 & 0.211 & 0.301 & 0.193 & 0.295 & 0.212 & 0.300 & 0.244 & 0.334 \\
		
		\midrule
		\multirow{5}{*}{\textbf{Exchange}} 
		& 96 & \textbf{0.083} & \textbf{0.202} & 0.089 & 0.208 & \underline{0.084} & \underline{0.203} & 0.098 & 0.232 & 0.102 & 0.230 & 0.088 & 0.209 & 0.087 & 0.206 & 0.090 & 0.211 & 0.107 & 0.234 & 0.088 & 0.218 & 0.139 & 0.265 \\
		& 192 & \textbf{0.168} & \textbf{0.292} & 0.183 & \underline{0.302} & 0.187 & 0.307 & 0.202 & 0.324 & 0.195 & 0.317 & 0.183 & 0.308 & 0.193 & 0.310 & 0.186 & 0.307 & 0.226 & 0.344 & \underline{0.176} & 0.315 & 0.241 & 0.375 \\
		& 336 & \underline{0.322} & \textbf{0.411} & 0.349 & 0.431 & 0.374 & 0.441 & 0.355 & 0.433 & 0.359 & 0.436 & 0.336 & \underline{0.418} & 0.345 & 0.425 & 0.339 & 0.424 & 0.367 & 0.448 & \textbf{0.313} & 0.427 & 0.392 & 0.468 \\
		& 720 & \textbf{0.838} & \textbf{0.684} & 0.880 & 0.707 & 0.972 & 0.739 & 0.896 & 0.718 & 0.940 & 0.738 & 0.893 & 0.714 & 1.008 & 0.747 & 0.898 & 0.718 & 0.964 & 0.746 & \underline{0.839} & \underline{0.695} & 1.110 & 0.802 \\
		& avg & \textbf{0.353} & \textbf{0.397} & 0.375 & \underline{0.412} & 0.404 & 0.423 & 0.388 & 0.427 & 0.399 & 0.430 & 0.375 & 0.412 & 0.408 & 0.422 & 0.378 & 0.415 & 0.416 & 0.443 & \underline{0.354} & 0.414 & 0.471 & 0.478 \\
		
		\midrule
		\multirow{5}{*}{\textbf{Solar-Energy}} 
		& 96 & \textbf{0.187} & \textbf{0.234} & 0.216 & 0.298 & 0.231 & 0.288 & 0.286 & 0.295 & 0.210 & 0.246 & \underline{0.203} & \underline{0.238} & 0.215 & 0.294 & 0.234 & 0.286 & 0.250 & 0.292 & 0.290 & 0.378 & 0.310 & 0.331 \\
		& 192 & \textbf{0.205} & \textbf{0.256} & 0.241 & 0.282 & 0.290 & 0.323 & 0.323 & 0.318 & 0.265 & 0.290 & \underline{0.233} & \underline{0.261} & 
		0.237 & 0.275 & 0.267 & 0.310 & 0.296 & 0.318 & 0.320 & 0.398 & 0.734 & 0.725 \\
		& 336 & \textbf{0.213} & \textbf{0.271} & 0.263 & 0.304 & 0.326 & 0.345 & 0.364 & 0.339 & 0.294 & 0.318 & \underline{0.248} & \underline{0.273} & 
		0.252 & 0.298 & 0.290 & 0.315 & 0.319 & 0.330 & 0.353 & 0.415 & 0.750 & 0.735 \\
		& 720 & \textbf{0.214} & \textbf{0.267} & 0.251 & 0.313 & 0.300 & 0.329 & 0.355 & 0.335 & 0.285 & 0.315 & 0.249 & \underline{0.276} & \underline{0.244} & 0.293 & 0.289 & 0.317 & 0.338 & 0.337 & 0.356 & 0.413 & 0.769 & 0.765 \\
		& avg & \textbf{0.205} & \textbf{0.257} & 0.243 & 0.299 & 0.287 & 0.321 & 0.332 & 0.322 & 0.263 & 0.292 & \underline{0.233} & \underline{0.262} & 
		0.237 & 0.290 & 0.270 & 0.307 & 0.301 & 0.319 & 0.330 & 0.401 & 0.641 & 0.639 \\
		\midrule
		\multirow{5}{*}{\textbf{Weather}}
		& 96 & 0.173 & 0.213 & 0.164 & \underline{0.210} & \underline{0.161} & \textbf{0.208} & 0.170 & 0.217 & 0.163 & 0.212 & 0.174 & 0.214 & 0.165 & 0.212 & 0.177 & 0.218 & 0.172 & 0.220 & 0.196 & 0.255 & \textbf{0.158} & 0.230 \\
		& 192 & 0.222 & 0.257 & 0.210 & \underline{0.251} & \underline{0.207} & \textbf{0.249} & 0.216 & 0.257 & 0.211 & 0.254 & 0.221 & 0.254 & 0.209 & 0.253 & 0.225 & 0.259 & 0.219 & 0.261 & 0.237 & 0.296 & \textbf{0.206} & 0.277 \\
		& 336 & 0.277 & 0.296 & 0.265 & \textbf{0.290} & \textbf{0.263} & \underline{0.290} & 0.270 & 0.294 & 0.273 & 0.299 & 0.278 & 0.296 & \underline{0.264} & 0.293 & 0.278 & 0.297 & 0.280 & 0.306 & 0.283 & 0.335 & 0.272 & 0.335 \\
		& 720 & 0.351 & 0.346 & 0.343 & \underline{0.342} & \textbf{0.340} & \textbf{0.341} & 0.348 & 0.344 & 0.351 & 0.348 & 0.358 & 0.347 & \underline{0.342} & 0.345 & 0.354 & 0.348 & 0.365 & 0.359 & 0.345 & 0.381 & 0.398 & 0.418 \\
		& avg & 0.256 & 0.278 & 0.246 & \underline{0.273} & \textbf{0.243} & \textbf{0.272} & 0.251 & 0.278 & 0.249 & 0.278 & 0.258 & 0.278 & \underline{0.245} & 0.276 & 0.259 & 0.281 & 0.259 & 0.287 & 0.265 & 0.317 & 0.259 & 0.315 \\
		
		\midrule
		\multirow{5}{*}{\textbf{Traffic}}
		& 96 & \textbf{0.392} & \textbf{0.263} & 0.511 & 0.324 & 0.608 & 0.383 & 0.576 & 0.342 & 0.608 & 0.349 & \underline{0.395} & \underline{0.268} & 0.462 & 0.285 & 0.544 & 0.359 & 0.593 & 0.321 & 0.650 & 0.396 & 0.522 & 0.290 \\
		& 192 & \textbf{0.415} & \textbf{0.273} & 0.529 & 0.330 & 0.571 & 0.364 & 0.596 & 0.361 & 0.634 & 0.371 & \underline{0.417} & \underline{0.276} & 
		0.473 & 0.296 & 0.540 & 0.354 & 0.617 & 0.336 & 0.598 & 0.370 & 0.530 & 0.293 \\
		& 336 & \textbf{0.430} & \textbf{0.279} & 0.545 & 0.334 & 0.561 & 0.364 & 0.630 & 0.371 & 0.669 & 0.388 & \underline{0.433} & \underline{0.283} & 
		0.498 & 0.296 & 0.551 & 0.358 & 0.629 & 0.336 & 0.605 & 0.373 & 0.558 & 0.305 \\
		& 720 & \textbf{0.463} & \textbf{0.297} & 0.580 & 0.351 & 0.619 & 0.385 & 0.667 & 0.390 & 0.729 & 0.420 & \underline{0.467} & \underline{0.302} & 
		0.506 & 0.313 & 0.586 & 0.375 & 0.640 & 0.350 & 0.645 & 0.394 & 0.589 & 0.328 \\
		& avg & \textbf{0.425} & \textbf{0.278} & 0.541 & 0.335 & 0.590 & 0.374 & 0.617 & 0.366 & 0.660 & 0.382 & \underline{0.428} & \underline{0.282} & 
		0.485 & 0.298 & 0.555 & 0.362 & 0.620 & 0.336 & 0.625 & 0.383 & 0.550 & 0.304 \\
		\midrule
		\multirow{5}{*}{\textbf{PEMS03}}
		& 12 & \textbf{0.065} & \textbf{0.168} & 0.077 & 0.185 & 0.095 & 0.202 & 0.091 & 0.204 & 0.078 & 0.187 & \underline{0.071} & \underline{0.174} & 0.076 & 0.185 & 0.073 & 0.178 & 0.085 & 0.192 & 0.122 & 0.243 & 0.090 & 0.203 \\
		& 24 & \textbf{0.086} & \textbf{0.195} & 0.119 & 0.231 & 0.138 & 0.221 & 0.113 & 0.213 & 0.108 & 0.218 & \underline{0.093} & \underline{0.201} & 
		0.112 & 0.224 & 0.105 & 0.212 & 0.118 & 0.223 & 0.201 & 0.317 & 0.121 & 0.240 \\
		& 48 & \textbf{0.123} & \textbf{0.235} & 0.184 & 0.291 & 0.180 & 0.279 & 0.153 & 0.269 & 0.178 & 0.272 & \underline{0.125} & \underline{0.236} & 
		0.181 & 0.281 & 0.159 & 0.264 & 0.155 & 0.260 & 0.333 & 0.425 & 0.202 & 0.317 \\
		& 96 & \underline{0.172} & \underline{0.277} & 0.251 & 0.336 & 0.272 & 0.329 & 0.232 & 0.324 & 0.238 & 0.328 & \textbf{0.164} & \textbf{0.275} & 
		0.205 & 0.343 & 0.210 & 0.305 & 0.228 & 0.317 & 0.457 & 0.515 & 0.262 & 0.367 \\
		& avg & \textbf{0.112} & \textbf{0.219} & 0.158 & 0.261 & 0.171 & 0.258 & 0.147 & 0.253 & 0.150 & 0.251 & \underline{0.113} & \underline{0.221} & 
		0.144 & 0.258 & 0.137 & 0.240 & 0.147 & 0.248 & 0.278 & 0.375 & 0.169 & 0.281 \\
		\midrule
		\multirow{5}{*}{\textbf{PEMS04}}
		& 12 & \textbf{0.076} & \textbf{0.178} & 0.096 & 0.205 & 0.102 & 0.202 & 0.092 & 0.208 & 0.086 & 0.199 & \underline{0.078} & \underline{0.183} & 0.091 & 0.203 & 0.085 & 0.189 & 0.087 & 0.195 & 0.148 & 0.272 & 0.098 & 0.218 \\
		& 24 & \textbf{0.091} & \textbf{0.199} & 0.128 & 0.247 & 0.133 & 0.241 & 0.112 & 0.253 & 0.101 & 0.218 & \underline{0.095} & \underline{0.205} & 
		0.129 & 0.243 & 0.115 & 0.222 & 0.103 & 0.215 & 0.224 & 0.340 & 0.131 & 0.256 \\
		& 48 & \textbf{0.119} & \textbf{0.231} & 0.164 & 0.309 & 0.142 & 0.271 & 0.128 & 0.247 & 0.127 & 0.247 & \underline{0.120} & \underline{0.233} & 
		0.199 & 0.303 & 0.167 & 0.273 & 0.136 & 0.250 & 0.355 & 0.437 & 0.205 & 0.326 \\
		& 96 & \textbf{0.150} & \textbf{0.262} & 0.221 & 0.355 & 0.214 & 0.321 & 0.163 & 0.286 & 0.174 & 0.292 & \underline{0.150} & \underline{0.262} & 
		0.223 & 0.338 & 0.211 & 0.310 & 0.190 & 0.303 & 0.452 & 0.504 & 0.402 & 0.457 \\
		& avg & \textbf{0.109} & \textbf{0.218} & 0.152 & 0.279 & 0.148 & 0.259 & 0.124 & 0.249 & 0.122 & 0.239 & \underline{0.111} & \underline{0.221} & 
		0.161 & 0.272 & 0.145 & 0.249 & 0.129 & 0.241 & 0.295 & 0.388 & 0.209 & 0.314 \\
		\midrule
		\multirow{5}{*}{\textbf{PEMS07}}
		& 12 & \textbf{0.058} & \textbf{0.153} & 0.072 & 0.174 & 0.083 & 0.196 & 0.073 & 0.186 & 0.079 & 0.182 & \underline{0.067} & 0.165 & 0.069 & 0.175 & 0.068 & \underline{0.163} & 0.082 & 0.181 & 0.115 & 0.242 & 0.094 & 0.200 \\
		& 24 & \textbf{0.076} & \textbf{0.176} & 0.117 & 0.222 & 0.101 & 0.211 & 0.110 & 0.209 & 0.099 & 0.206 & \underline{0.088} & \underline{0.190} & 
		0.107 & 0.216 & 0.102 & 0.201 & 0.101 & 0.204 & 0.210 & 0.329 & 0.139 & 0.247 \\
		& 48 & \textbf{0.106} & \textbf{0.211} & 0.169 & 0.246 & 0.172 & 0.264 & 0.165 & 0.264 & 0.133 & 0.239 & \underline{0.110} & \underline{0.215} & 
		0.175 & 0.272 & 0.170 & 0.261 & 0.134 & 0.238 & 0.398 & 0.458 & 0.311 & 0.369 \\
		& 96 & \textbf{0.132} & \textbf{0.241} & 0.192 & 0.291 & 0.199 & 0.289 & 0.268 & 0.328 & 0.179 & 0.279 & \underline{0.139} & \underline{0.245} & 
		0.295 & 0.350 & 0.236 & 0.308 & 0.181 & 0.279 & 0.594 & 0.553 & 0.396 & 0.442 \\
		& avg & \textbf{0.093} & \textbf{0.195} & 0.138 & 0.233 & 0.139 & 0.240 & 0.154 & 0.247 & 0.122 & 0.227 & \underline{0.101} & \underline{0.204} & 
		0.162 & 0.253 & 0.144 & 0.233 & 0.124 & 0.225 & 0.329 & 0.395 & 0.235 & 0.315 \\
		\midrule
		\multirow{5}{*}{\textbf{PEMS08}}
		& 12 & \textbf{0.076} & \textbf{0.175} & 0.088 & 0.191 & 0.103 & 0.217 & 0.095 & 0.202 & 0.105 & 0.211 & \underline{0.079} & \underline{0.182} & 0.093 & 0.202 & 0.098 & 0.205 & 0.112 & 0.212 & 0.154 & 0.276 & 0.165 & 0.214 \\
		& 24 & \textbf{0.106} & \textbf{0.206} & 0.139 & 0.243 & 0.176 & 0.287 & 0.118 & 0.231 & 0.141 & 0.243 & \underline{0.115} & \underline{0.219} & 
		0.148 & 0.261 & 0.162 & 0.266 & 0.141 & 0.238 & 0.248 & 0.353 & 0.215 & 0.260 \\
		& 48 & \textbf{0.183} & \underline{0.251} & 0.265 & 0.302 & 0.241 & 0.329 & 0.189 & 0.285 & 0.211 & 0.300 & \underline{0.186} & \textbf{0.235} & 
		0.280 & 0.357 & 0.238 & 0.311 & 0.198 & 0.283 & 0.440 & 0.470 & 0.315 & 0.355 \\
		& 96 & \textbf{0.215} & \textbf{0.262} & 0.363 & 0.336 & 0.332 & 0.331 & 0.302 & 0.301 & 0.364 & 0.387 & \underline{0.221} & \underline{0.267} & 
		0.301 & 0.363 & 0.303 & 0.318 & 0.320 & 0.351 & 0.674 & 0.565 & 0.377 & 0.397 \\
		& avg & \textbf{0.145} & \textbf{0.224} & 0.214 & 0.268 & 0.213 & 0.291 & 0.176 & 0.255 & 0.205 & 0.285 & \underline{0.150} & \underline{0.226} & 
		0.206 & 0.296 & 0.200 & 0.275 & 0.193 & 0.271 & 0.379 & 0.416 & 0.268 & 0.307 \\
		\bottomrule[1.2pt]
	\end{tabular}
	
	\label{tab:amse}
\end{table*}

\begin{table}[t]
	\centering
	\small
	\renewcommand{\arraystretch}{1.}
	\caption{Overall forecasting performance of Time-TK and TimeKAN on 10 benchmark datasets. We report mean $\pm$ standard deviation over three runs.}
	\label{tab:overall_performance}
	\setlength{\tabcolsep}{3pt}
	\begin{tabular}{lcccc}
		\toprule
		\multirow{2}{*}{Datasets} & \multicolumn{2}{c}{Time-TK} & \multicolumn{2}{c}{TimeKAN} \\
		\cmidrule(lr){2-3} \cmidrule(lr){4-5}
		& MSE & MAE & MSE & MAE \\
		\midrule
		ETTh1        & $0.433\pm0.005$ & $0.430\pm0.003$ & $0.425\pm0.005$ & $0.430\pm0.004$ \\
		ETTh2        & $0.371\pm0.004$ & $0.399\pm0.005$ & $0.389\pm0.002$ & $0.408\pm0.003$ \\
		ETTm1        & $0.380\pm0.004$ & $0.395\pm0.004$ & $0.381\pm0.005$ & $0.397\pm0.004$ \\
		ETTm2        & $0.276\pm0.002$ & $0.321\pm0.003$ & $0.281\pm0.003$ & $0.327\pm0.003$ \\
		Electricity  & $0.175\pm0.005$ & $0.270\pm0.004$ & $0.198\pm0.003$ & $0.288\pm0.002$ \\
		Solar-Energy & $0.203\pm0.006$ & $0.265\pm0.003$ & $0.278\pm0.005$ & $0.315\pm0.004$ \\
		Weather      & $0.255\pm0.003$ & $0.278\pm0.002$ & $0.244\pm0.005$ & $0.273\pm0.003$ \\
		Traffic      & $0.425\pm0.004$ & $0.278\pm0.002$ & $0.593\pm0.003$ & $0.378\pm0.005$ \\
		PEMS04       & $0.109\pm0.005$ & $0.217\pm0.004$ & $0.157\pm0.007$ & $0.263\pm0.008$ \\
		PEMS08       & $0.149\pm0.005$ & $0.232\pm0.006$ & $0.217\pm0.004$ & $0.293\pm0.006$ \\
		\bottomrule
	\end{tabular}
\end{table}
\subsection{Evaluation metrics}\label{apA.3}
Mean square error (MSE) and Mean absolute error (MAE) are commonly used as measures of forecasting performance in time series forecasting tasks. MSE represents the average of the squared differences between the predicted and actual values, giving more weight to larger deviations. MAE reflects the average of the absolute differences, thus providing a more balanced picture of the overall magnitude of the error. Together, these metrics constitute a comprehensive assessment of model accuracy. The mathematical definitions are as follows:
\begin{equation}\label{eq:1}
	\begin{aligned}
		\begin{array}{l}
			MSE = \frac{1}{\mathcal{F}}\sum\limits_{i = 1}^\mathcal{F} {{{({\mathcal{Y}_i} - {{\mathcal{ \hat Y} }_i})}^2}} \\
			MAE = \frac{1}{\mathcal{F}}\sum\limits_{i = 1}^\mathcal{F} {\left| {{\mathcal{Y}_i} - {{\mathcal{\hat Y} }_i}} \right|} \\
			RMSE = \sqrt {\frac{1}{\mathcal{F}}\sum\limits_{i = 1}^\mathcal{F} {{{({\mathcal{Y}_i} - {{\mathcal{\hat Y}}_i})}^2}} } \\
			RSE = \frac{{\sqrt {\sum\limits_{i = 1}^\mathcal{F} {{{({\mathcal{Y}_i} - {{\mathcal{\hat Y}}_i})}^2}} } }}{{\sqrt {\sum\limits_{i = 1}^\mathcal{F} {{{({\mathcal{Y}_i} - \bar {\mathcal{Y}})}^2}} } }} \\
			MAPE = \frac{1}{\mathcal{F}}\sum\limits_{i = 1}^\mathcal{F} {\left| {\frac{{{\mathcal{Y}_i} - {{\mathcal{\hat Y}}_i}}}{{{\mathcal{Y}_i}}}} \right|} 
		\end{array}
	\end{aligned}
\end{equation}

Where \(\mathcal{F}\) represents the size of the lookback window, \(\mathcal{Y}_i\) represents the true value, and \({{{\mathcal{\hat Y} }_i}}\) represents the predicted value of the model.
\subsection{Experimental Details}\label{apA.2}

Specifically, all experiments in this study were implemented using PyTorch and executed on a single NVIDIA A100-SXM GPU with 40 GB of memory. The models were trained using the Adam optimizer \citep{kingma2014adam} and optimized with the L2 loss function. The training-validation-test split follows existing practices in the literature, such as those reported for iTransformer \citep{15} and TimesNet \citep{28}. Specifically, the ETT and PEMS series datasets were split in a 6:2:2 ratio, while the remaining datasets used a 7:1:2 split. Time-TK was trained for 30 epochs with early stopping based on a patience of 3 on the validation set. For most datasets, the learning rate was set to 0.003, except for smaller datasets such as the ETT series, where it was reduced to 0.002. The batch size was adjusted based on the dataset size to maximize GPU utilization while avoiding memory overflow. For example, a batch size of 16 was used for the Traffic dataset, while a batch size of 64 was used for the Weather dataset. To ensure reproducibility, all experiments were conducted with a fixed random seed of 2024.

%
%
%

\begin{figure*}[!ht]     
	\centering
	\begin{minipage}[b]{0.24\textwidth}
		\includegraphics[width=\textwidth]{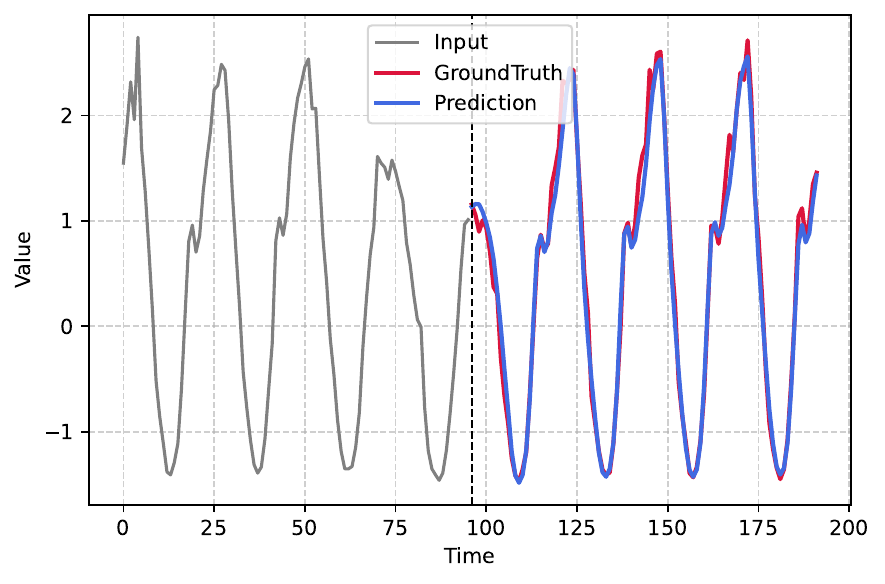}
		\centering
		\parbox{\textwidth}{\centering (Time-TK)}
	\end{minipage}
	\begin{minipage}[b]{0.24\textwidth}
		\includegraphics[width=\textwidth]{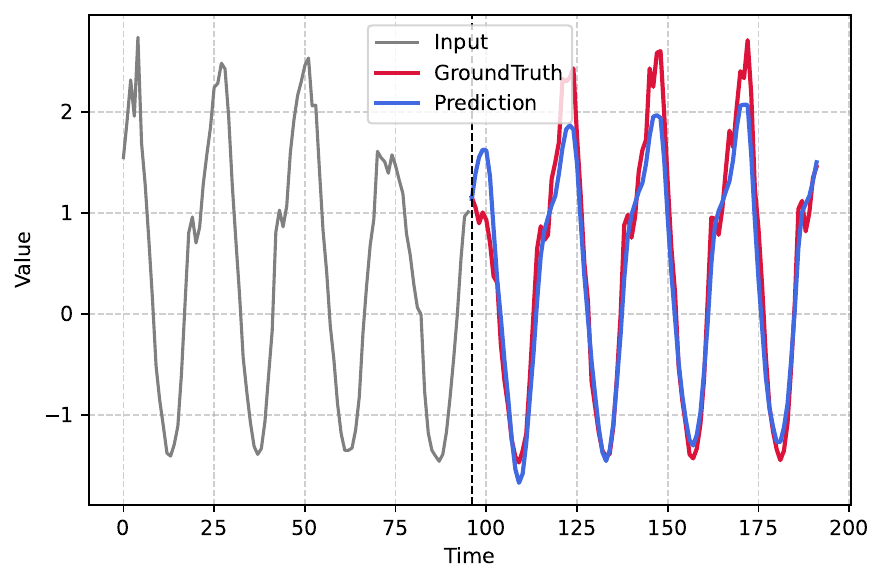}
		\centering
		\parbox{\textwidth}{\centering (TimeKAN)}
	\end{minipage}
	\begin{minipage}[b]{0.24\textwidth}
		\includegraphics[width=\textwidth]{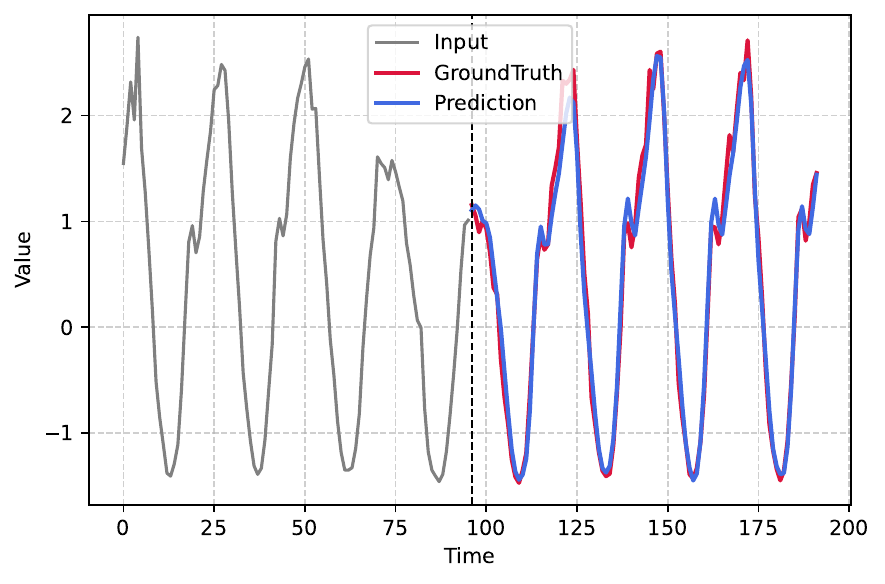}
		\centering
		\parbox{\textwidth}{\centering (iTransformer)}
	\end{minipage}
	\begin{minipage}[b]{0.24\textwidth}
		\includegraphics[width=\textwidth]{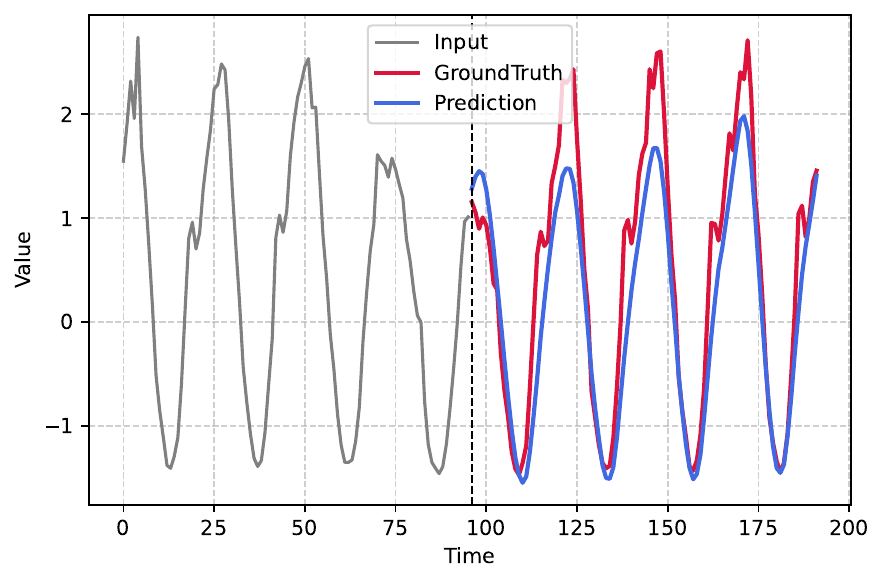}
		\centering
		\parbox{\textwidth}{\centering (PatchTST)}
	\end{minipage}
	\caption{The performance of each model is visualized and compared on the traffic dataset with lookback window \({\mathcal L}\) =96, prediction window \({\mathcal F}\) = 96.}
	\label{fig:96}
\end{figure*}

\begin{figure*}[!ht]     
	\centering
	\begin{minipage}[b]{0.24\textwidth}
		\includegraphics[width=\textwidth]{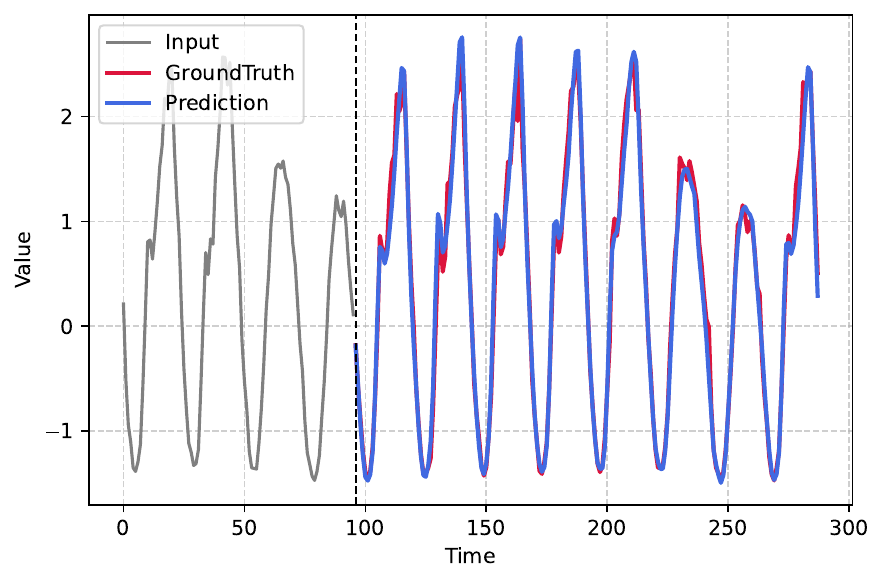}
		\centering
		\parbox{\textwidth}{\centering (Time-TK)}
	\end{minipage}
	\begin{minipage}[b]{0.24\textwidth}
		\includegraphics[width=\textwidth]{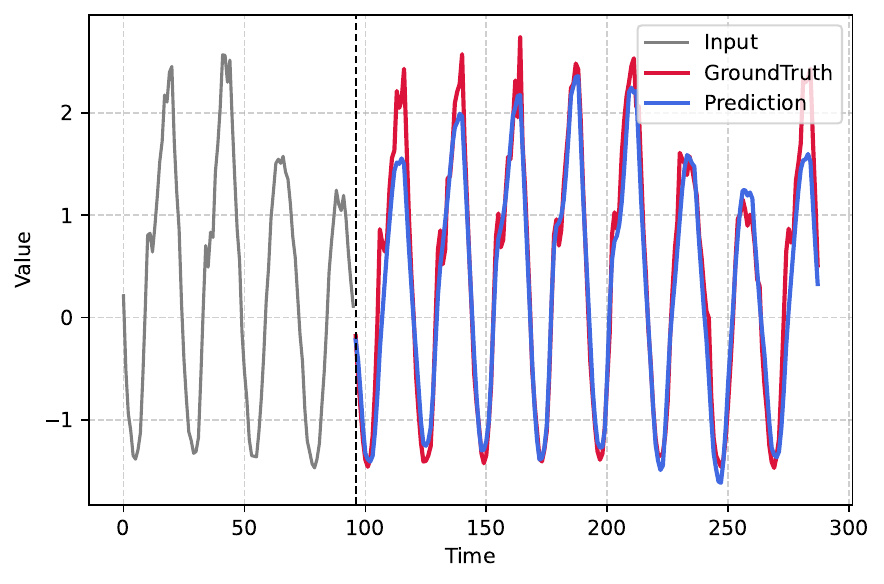}
		\centering
		\parbox{\textwidth}{\centering (TimeKAN)}
	\end{minipage}
	\begin{minipage}[b]{0.24\textwidth}
		\includegraphics[width=\textwidth]{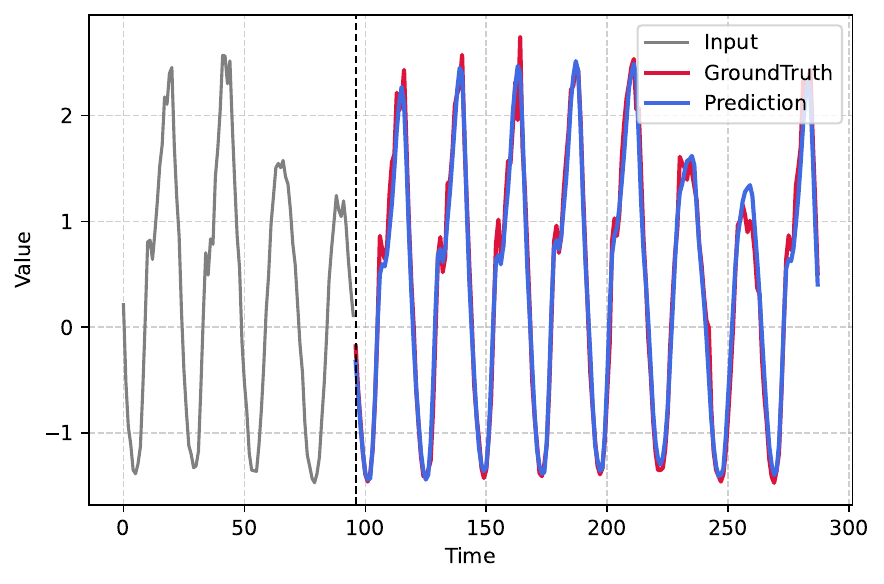}
		\centering
		\parbox{\textwidth}{\centering (iTransformer)}
	\end{minipage}
	\begin{minipage}[b]{0.24\textwidth}
		\includegraphics[width=\textwidth]{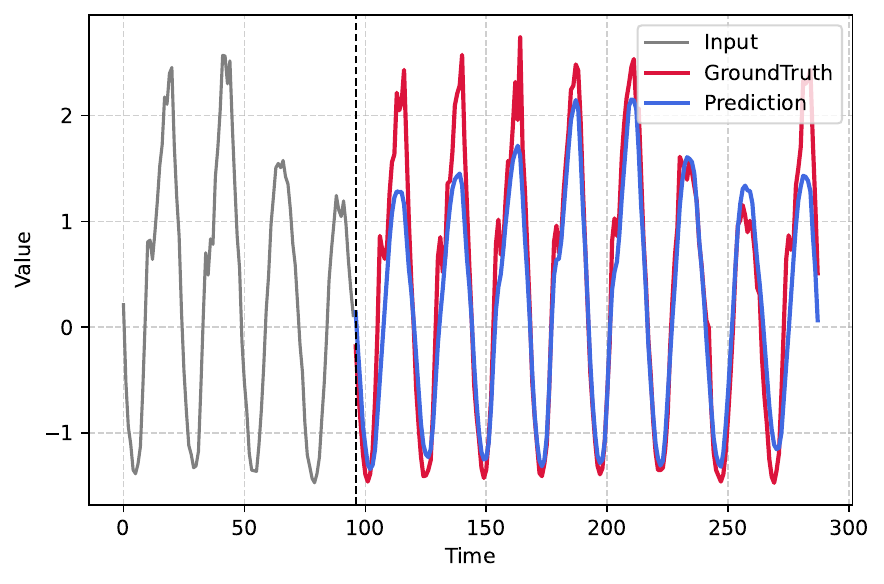}
		\centering
		\parbox{\textwidth}{\centering (PatchTST)}
	\end{minipage}
	\caption{The performance of each model is visualized and compared on the traffic dataset with lookback window \({\mathcal L}\) =96, prediction window \({\mathcal F}\) = 192.}
	\label{fig:192}
\end{figure*}

\begin{figure*}[!ht]     
	\centering
	\begin{minipage}[b]{0.24\textwidth}
		\includegraphics[width=\textwidth]{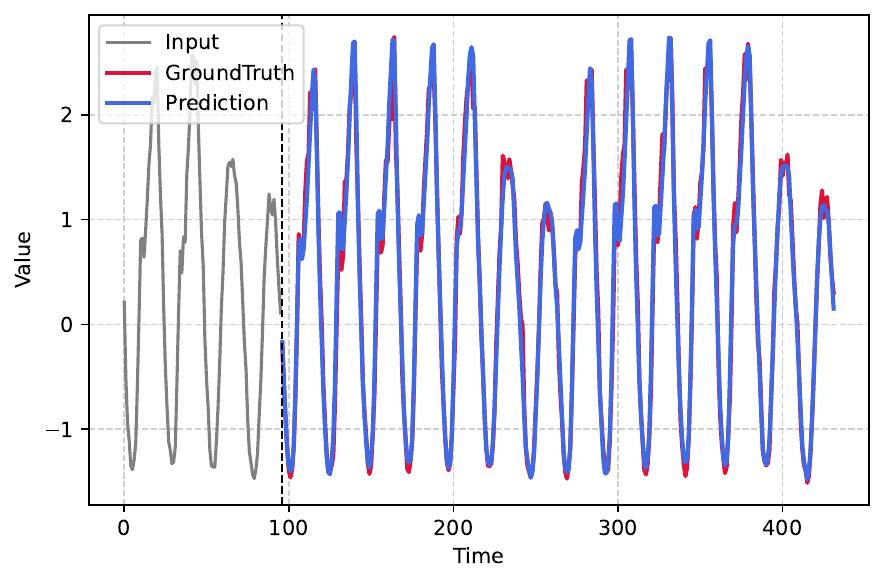}
		\centering
		\parbox{\textwidth}{\centering (Time-TK)}
	\end{minipage}
	\begin{minipage}[b]{0.24\textwidth}
		\includegraphics[width=\textwidth]{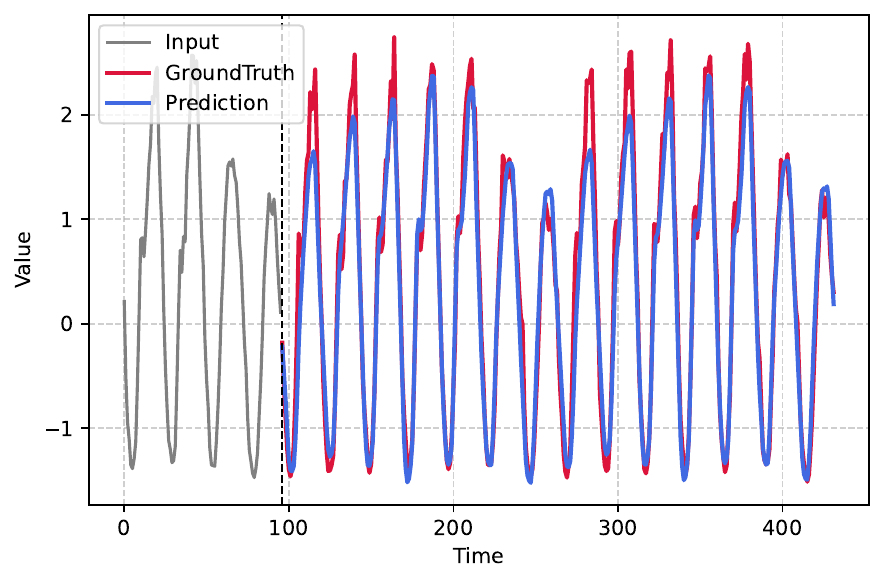}
		\centering
		\parbox{\textwidth}{\centering (TimeKAN)}
	\end{minipage}
	\begin{minipage}[b]{0.24\textwidth}
		\includegraphics[width=\textwidth]{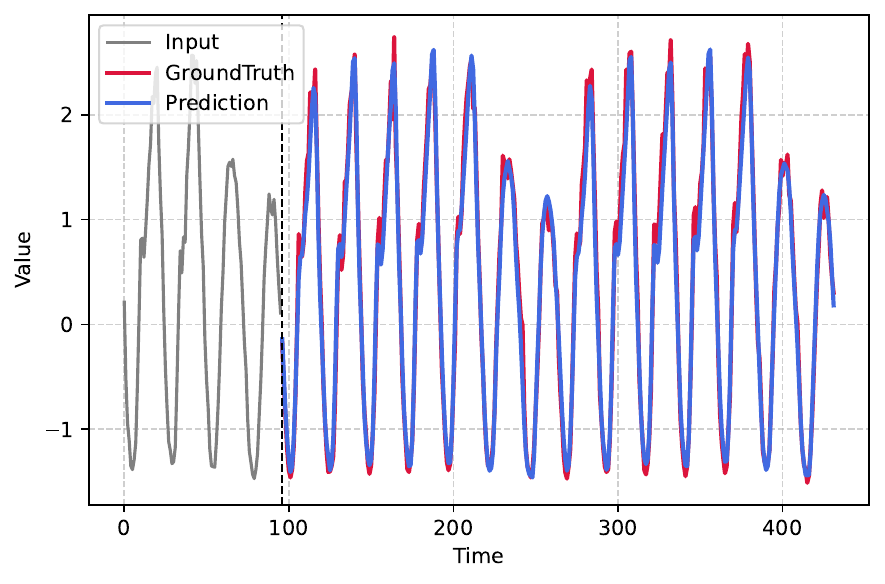}
		\centering
		\parbox{\textwidth}{\centering (iTransformer)}
	\end{minipage}
	\begin{minipage}[b]{0.24\textwidth}
		\includegraphics[width=\textwidth]{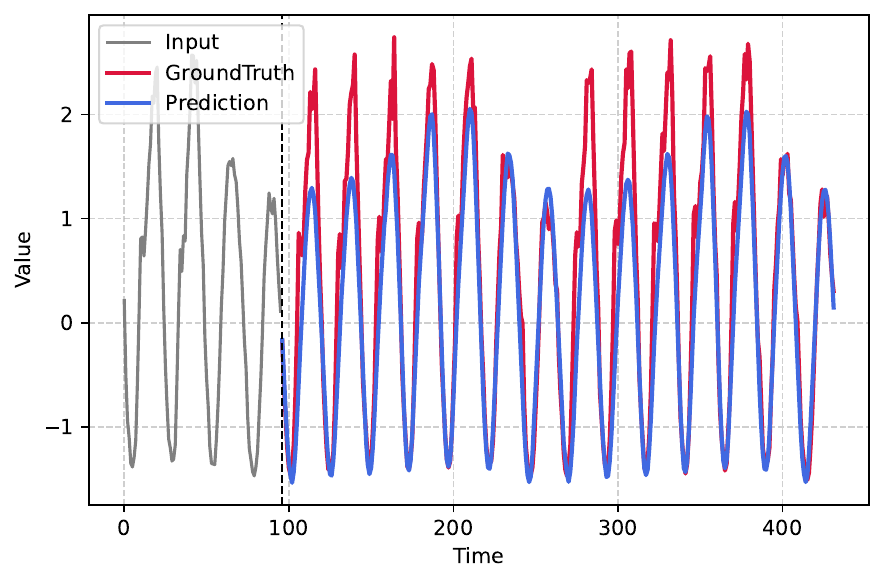}
		\centering
		\parbox{\textwidth}{\centering (PatchTST)}
	\end{minipage}
	\caption{The performance of each model is visualized and compared on the traffic dataset with lookback window \({\mathcal L}\) =96, prediction window \({\mathcal F}\) = 336.}
	\label{fig:336}
\end{figure*}
\begin{figure*}[!ht]     
	\centering
	\begin{minipage}[b]{0.24\textwidth}
		\includegraphics[width=\textwidth]{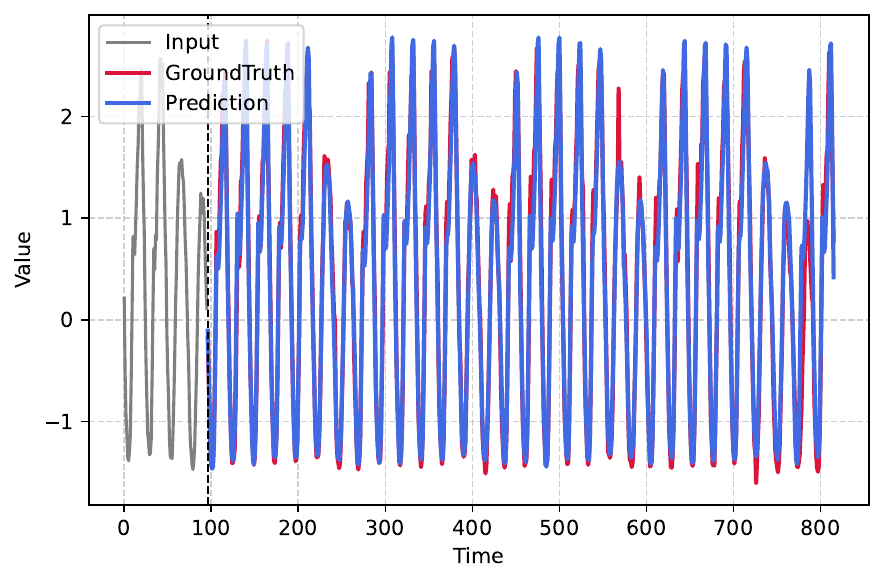}
		\centering
		\parbox{\textwidth}{\centering (Time-TK)}
	\end{minipage}
	\begin{minipage}[b]{0.24\textwidth}
		\includegraphics[width=\textwidth]{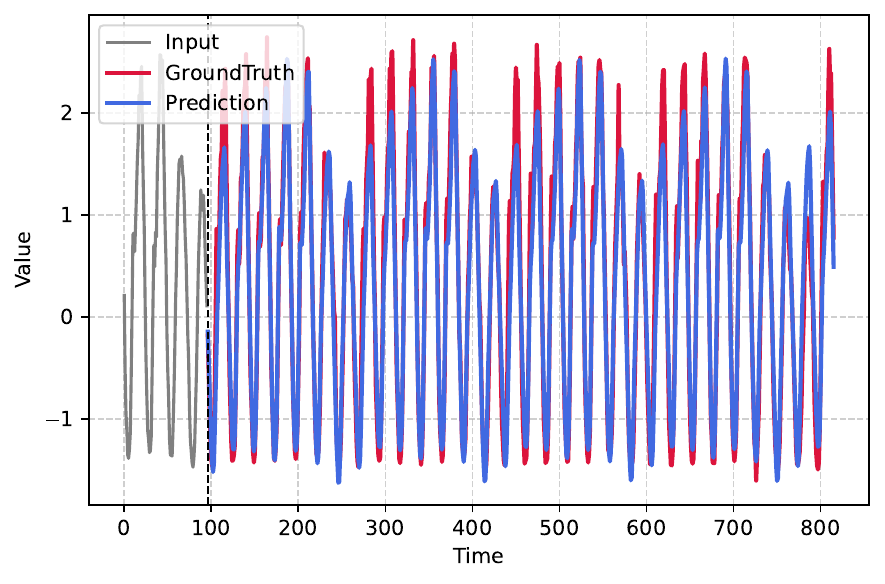}
		\centering
		\parbox{\textwidth}{\centering (TimeKAN)}
	\end{minipage}
	\begin{minipage}[b]{0.24\textwidth}
		\includegraphics[width=\textwidth]{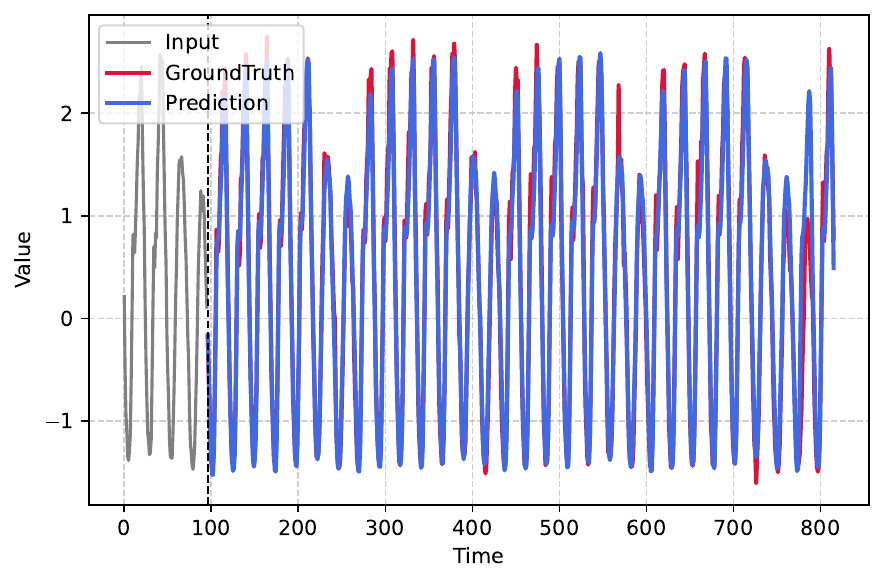}
		\centering
		\parbox{\textwidth}{\centering (iTransformer)}
	\end{minipage}
	\begin{minipage}[b]{0.24\textwidth}
		\includegraphics[width=\textwidth]{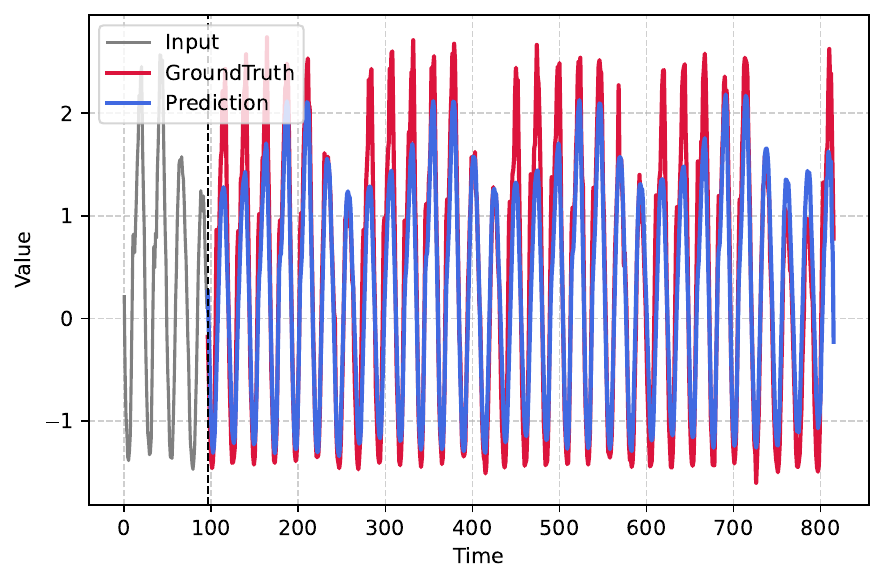}
		\centering
		\parbox{\textwidth}{\centering (PatchTST)}
	\end{minipage}
	\caption{The performance of each model is visualized and compared on the traffic dataset with lookback window \({\mathcal L}\) =96, prediction window \({\mathcal F}\) = 720.}
	\label{fig:720}
\end{figure*}

%

\section{Full Results}\label{apB}
In this section, we present the full results of the experiments conducted in this study, as shown in Table \ref{tab:amse}. The results cover evaluations across different datasets, demonstrating the robustness of the proposed method in various real-world scenarios. Additionally, we provide detailed performance curves, as shown in Figures 1-4, for a comprehensive understanding of the model's behavior. This section aims to provide readers with a complete view of the experimental outcomes, complementing the discussions in the main body of the paper. We provide detailed tables and figures to facilitate a deeper interpretation of the results and to support the conclusions drawn in the preceding sections. Table \ref{tab:overall_performance} shows the statistical significance test results.
\section{More Details of Time-TK}\label{apC}
This algorithm describes the basic process of the Time-TK model. First, the input time series \({\mathcal X} \in \mathbb{R}^{N \times \mathcal L}\) is normalized using RevIN, resulting in \({\mathcal X}_n\). Then, the Multi-Offset Temporal Embedding (MOTE) method is applied to divide the normalized data into multiple subsequences with different time offsets,\(\{ {\mathcal{M}_1},...,{M_{\mathcal{O}}}\}\). These subsequences are further processed by the Multi-Offset Interactive KAN (MI-KAN) module, yielding \(\{ {\mathcal{M}'_1},...,{M'_{\mathcal{O}}} \}\). For each subsequence, a Multi-Head Self-Attention (MSA) mechanism is applied to capture interactions, resulting in \({\mathcal{A}_i}\). Subsequently, these interaction results are fused with the original sequence \({\mathcal X}\) through a global Multi-Head Self-Attention operation to generate the final representation  \(	\mathcal{H}\). Finally, the predictor is applied to \(	\mathcal{H}\) to obtain the prediction \(\hat{\mathcal Y}\), and the result is denormalized via the inverse ReVIN to obtain the final prediction output \(\hat{\mathcal Y}\). The algorithm continues until the stopping criteria are met.

\end{document}